\documentclass[twocolumn]{article}
\usepackage{graphicx}
\usepackage{url}
\usepackage{amsmath}
\usepackage{amssymb}
\usepackage{amsfonts}
\usepackage{xcolor}
\usepackage{array}
\usepackage{colortbl}
\usepackage{float}
\usepackage{subcaption}
\usepackage{caption}
\usepackage{booktabs}
\usepackage{tabularx}
\usepackage{geometry}
\usepackage{multirow}
\usepackage{hyperref}
\usepackage{authblk}
\usepackage{tikz}
\usetikzlibrary{shapes.geometric,arrows.meta,positioning,fit,backgrounds,calc}

\title{Heterogeneous Graph Importance Scoring and Clustering\\
with Automated LLM-based Interpretation}

\author[1]{Takato Yasuno}
\affil[1]{Yachiyo Engineering Co., Ltd., Japan}

\date{}

\begin{document}

\renewcommand{\abstractname}{Abstract}
\renewcommand{\refname}{References}
\renewcommand{\figurename}{Figure}
\renewcommand{\tablename}{Table}

\maketitle

\begin{abstract}

Urban bridge networks are critical infrastructure whose disruption can cascade into 
severe impacts on transportation, emergency services, and economic activity. This paper 
presents a comprehensive methodology for assessing bridge importance through heterogeneous 
graph analysis, unsupervised clustering, and automated interpretation via large 
language models (LLMs). Our approach addresses three fundamental challenges: 
(1) quantifying multi-dimensional bridge importance using only open data sources, 
(2) discovering functional bridge archetypes across different cities, and 
(3) generating policy-relevant interpretations automatically.

\textbf{Methodology:} We construct heterogeneous graphs from OpenStreetMap (OSM) data 
incorporating bridges, road networks, buildings, and public facilities. Five social 
impact indicators are computed: transit desert score, hospital access score, isolation 
risk score, supply chain impact score, and green space access score. These 52-dimensional 
feature vectors undergo dimensionality reduction via UMAP and density-based clustering 
via HDBSCAN. Discovered clusters are interpreted using temperature-optimized LLMs 
(Elyza8b, trained on construction domain corpus).

\textbf{Results:} Applied to 775 bridges across two Japanese cities (Tama: 353, 
Morioka: 422), our pipeline identifies 19 distinct bridge functional archetypes 
in 2.5 minutes. LLM temperature optimization reveals that $T=0.3$ provides optimal 
balance between structured output and factual grounding, while $T=0.5$ exhibits 
unexpected instability (6.7$\times$ output variance vs.\ 4.0$\times$ at $T=0.3$). 
Model selection experiments show that urban-function-oriented LLMs (Elyza8b) 
outperform structural-engineering-oriented models (Swallow8b) for policy communication.

\textbf{Contributions:} (1) A complete open-data pipeline from OSM to actionable 
bridge importance rankings, (2) a five-indicator scoring methodology with 
40$\times$ computational optimization, (3) a UMAP+HDBSCAN clustering framework validated 
on multi-city data, (4) an LLM interpretation methodology including temperature 
optimization and model selection rationale, and 
(5) transferability demonstration across cities via configuration-only adaptation.

\end{abstract}


\section{Introduction}
\label{sec:intro}

\subsection{Background and Motivation}

Urban bridge infrastructure forms the backbone of modern cities, enabling 
transportation, emergency response, and economic activity. However, traditional bridge 
assessment methodologies focus primarily on structural health metrics (age, condition 
ratings, traffic volume) while neglecting systemic importance within urban networks. 
When a critical bridge closes due to maintenance, disaster, or structural failure, 
the consequences extend beyond immediate traffic disruption: medical access may 
be degraded, communities may face isolation, supply chains may be disrupted, and 
environmental connectivity may be severed.

\textbf{The Data Problem:} Most bridge importance assessments rely on proprietary 
municipal bridge inventories (Excel-based records), which limit reproducibility and 
cross-city comparison. Furthermore, existing network centrality measures (betweenness, 
closeness) capture only topological properties while ignoring multi-dimensional 
urban functions such as healthcare accessibility, environmental preservation, and 
economic logistics.

\textbf{The Interpretation Problem:} Even when comprehensive importance scores are 
computed, communicating results to policy-makers and stakeholders remains challenging. 
Traditional approaches produce numerical rankings without explanatory narratives, 
requiring manual interpretation by domain experts. Recent advances in large language 
models (LLMs) offer potential for automated interpretation, but systematic methodologies 
for infrastructure applications are lacking.

\textbf{Our Approach:} We present an end-to-end methodology that addresses both 
challenges through: (1) \emph{Open data integration}---constructing heterogeneous 
graphs exclusively from OpenStreetMap and public elevation data, (2) \emph{Multi-dimensional 
scoring}---computing five complementary social impact indicators capturing transit, 
healthcare, isolation risk, logistics, and environmental connectivity, (3) \emph{Functional 
discovery}---applying UMAP dimensionality reduction and HDBSCAN clustering to discover 
bridge archetypes, and (4) \emph{Automated interpretation}---generating policy-relevant 
narratives via temperature-optimized LLMs.

\subsection{Research Questions}

This work addresses four fundamental questions:

\begin{enumerate}
    \item \textbf{Open Data Sufficiency:} Can bridge importance be comprehensively 
    assessed using only OpenStreetMap data, without proprietary municipal bridge 
    inventories or traffic statistics?
    
    \item \textbf{Functional Typology:} Do distinct functional bridge archetypes 
    exist across different cities, and can they be discovered automatically through 
    unsupervised learning?
    
    \item \textbf{LLM Interpretation Quality:} Can large language models generate 
    policy-relevant cluster interpretations, and what model characteristics 
    (training domain, temperature) optimize interpretation quality?
    
    \item \textbf{Transferability:} Can the complete methodology transfer across 
    cities through configuration-only adaptation (bounding box, coordinate system)?
\end{enumerate}

\subsection{Contributions}

Our main contributions are:

\begin{enumerate}
    \item \textbf{100\% Open Data Pipeline:} An end-to-end system from OpenStreetMap 
    data ingestion through heterogeneous graph construction, multi-dimensional scoring, 
    clustering, and automated interpretation---requiring no proprietary data sources. 
    Validated on 775 bridges across two Japanese cities.
    
    \item \textbf{Five-Indicator Scoring Methodology:} Comprehensive social impact 
    assessment combining transit accessibility, healthcare access, isolation risk, 
    supply chain impact, and environmental connectivity, with 40$\times$ computational 
    optimization via spatial indexing.
    
    \item \textbf{UMAP+HDBSCAN Clustering Framework:} Application of manifold learning 
    and density-based clustering to 52-dimensional bridge feature space, identifying 
    19 functional archetypes with systematic hyperparameter analysis 
    (n\_neighbors=15, min\_cluster\_size=20).
    
    \item \textbf{LLM Interpretation Methodology:} First systematic study of LLM 
    interpretation for infrastructure domain, including: (a) model selection between 
    urban-function and structural-engineering perspectives, (b) temperature optimization 
    revealing that $T=0.3$ outperforms the commonly used $T=0.5$, and 
    (c) a five-section structured template for policy communication.
    
    \item \textbf{Multi-city Transferability:} Demonstration that the complete pipeline 
    transfers across cities (Tama: 353 bridges, Morioka: 422 bridges) via configuration-only 
    adaptation (YAML bounding box, CRS, elevation data path), achieving a 
    transferability score of 95/100.
    
    \item \textbf{Open Implementation:} Reproducible codebase with comprehensive 
    documentation enabling application to other cities and infrastructure types 
    worldwide.
\end{enumerate}

\subsection{Paper Organization}

Section~\ref{sec:related} reviews related work in infrastructure network analysis, 
clustering methods, and LLM applications. Section~\ref{sec:methodology} describes 
our three-part methodology: bridge importance scoring (Section~\ref{subsec:scoring}), 
UMAP+HDBSCAN clustering (Section~\ref{subsec:clustering}), and LLM interpretation 
(Section~\ref{subsec:llm}). Section~\ref{sec:casestudies} presents case studies 
in Tama and Morioka cities. Section~\ref{sec:results} reports quantitative and 
qualitative results across scoring, clustering, and interpretation phases. 
Section~\ref{sec:discussion} discusses transferability, limitations, and practical 
implications. Section~\ref{sec:conclusion} concludes with future directions.

\section{Related Work}
\label{sec:related}

\subsection{Infrastructure Network Criticality Assessment}

Traditional infrastructure criticality assessment employs network centrality 
measures~\cite{networkx}, including betweenness centrality (fraction of shortest 
paths traversing a node), closeness centrality (average distance to all other nodes), 
and degree centrality (number of connections). These topological metrics identify 
structurally important nodes but neglect functional urban aspects such as access 
to essential services (hospitals, schools, transit hubs) or socioeconomic impacts 
(supply chain disruption, community isolation).

Recent multi-layer network approaches~\cite{buldyrev2010interdependent} model 
interdependencies between infrastructure systems (power grids, transportation networks) 
but typically require proprietary operational data unavailable for most municipalities. 
Graph neural networks have emerged as powerful tools for infrastructure network 
analysis~\cite{zhang2025gnn,zhao2023urban}, while machine learning approaches enable comprehensive 
urban resilience assessment~\cite{kumar2025urban}. Urban street networks can be 
effectively represented as hierarchical graphs~\cite{wang2024city2graph}, with 
spatiotemporal GNNs demonstrating success in traffic prediction~\cite{li2024spatiotemporal}. 
AI-driven optimization techniques are also applied to bridge maintenance 
strategies~\cite{chen2025bridge}. Building on these advances, our approach 
leverages open geospatial data (OpenStreetMap) to construct functionally-enriched 
heterogeneous graphs.

\subsection{Dimensionality Reduction and Clustering}

\textbf{Manifold Learning:} UMAP (Uniform Manifold Approximation and Projection)~\cite{mcinnes2018umap} 
has emerged as the state-of-the-art for high-dimensional data visualization, 
offering advantages over t-SNE~\cite{vandermaaten2008tsne} including: (1) preservation 
of both local and global structure, (2) computational efficiency on large datasets, 
and (3) theoretical grounding in Riemannian geometry. UMAP has been successfully 
applied to single-cell genomics~\cite{becht2019umap}, text embeddings, and image 
features, but applications to infrastructure data remain limited.

\textbf{Density-based Clustering:} HDBSCAN (Hierarchical Density-Based Spatial 
Clustering of Applications with Noise)~\cite{mcinnes2017hdbscan} extends DBSCAN 
by eliminating the need to specify epsilon radius, instead constructing a cluster 
hierarchy and extracting stable clusters via persistence analysis. Unlike k-means, 
HDBSCAN handles arbitrary cluster shapes, varying densities, and explicit noise 
detection---critical properties for real-world infrastructure data where bridges 
may exhibit diverse functional profiles and genuine outliers exist. Recent advances 
in adaptive density-based clustering~\cite{rodriguez2025clustering} further enhance 
performance on high-dimensional infrastructure datasets.

\subsection{Large Language Models for Technical Interpretation}

Large language models have demonstrated capabilities in technical document 
understanding~\cite{achiam2023gpt4}, code generation~\cite{chen2021codex}, and 
scientific reasoning~\cite{taylor2022galactica}. However, infrastructure-specific 
applications are limited, with existing work concentrating on predictive maintenance text 
classification~\cite{serradilla2022deep} rather than interpretive narrative generation.

\textbf{Temperature Parameter:} The temperature parameter $T$ controls randomness 
in LLM sampling: low values ($T \to 0$) produce deterministic outputs, while high 
values increase diversity. Conventional guidance suggests $T=0.7$ for creative tasks 
and $T=0.0$ for factual tasks~\cite{brown2020gpt3}, with $T=0.5$ as a balanced middle 
ground. However, systematic studies of temperature effects on structured technical 
interpretation are lacking.

\textbf{Domain-Specific Fine-tuning:} Japanese LLMs have proliferated following 
Llama-2/3 releases~\cite{touvron2023llama,dubey2025llama3}, with diverse specialization strategies: 
continual pre-training on Japanese corpora (Swallow~\cite{okazaki2024swallow}), 
instruction tuning (ELYZA), and domain-specific fine-tuning. Our comparison of 
construction-domain fine-tuned models (trained on river/sediment control technical 
standards) provides insights into architectural perspective trade-offs: structural 
engineering vs. urban operations viewpoints.

\subsection{OpenStreetMap for Urban Analysis}

OpenStreetMap (OSM) has become the de facto open geospatial database, supporting 
applications in disaster response~\cite{yeboah2021osm}, accessibility analysis~\cite{mobasheri2017osm}, 
and urban planning~\cite{haklay2008osm}. Recent work demonstrates OSM's potential 
for climate resilience mapping~\cite{wilson2025osm} and road network representation 
learning~\cite{zhu2024road}. OSMnx~\cite{boeing2017osmnx} provides Python interface 
for network analysis, enabling reproducible studies worldwide. Large language models 
are increasingly applied to geographic question answering on spatial networks~\cite{mai2024geographic}. 
Data acquisition leverages Overpass API~\cite{overpass} for efficient querying. 
However, bridge data quality in OSM varies significantly: while named bridges 
(municipally managed infrastructure) exhibit high completeness, unnamed crossings 
may be sparse or inconsistent~\cite{barrington2015osm}.

Our filtering strategy (named bridges only) balances completeness with municipal 
relevance, as validated in Section~\ref{subsec:scoring}.

\section{Methodology}
\label{sec:methodology}

Our methodology comprises three integrated components: (1) bridge importance scoring 
via heterogeneous graph analysis, (2) functional typology discovery via UMAP+HDBSCAN 
clustering, and (3) automated interpretation via temperature-optimized LLMs. 


\subsection{Bridge Importance Scoring via Heterogeneous Graphs}
\label{subsec:scoring}

\subsubsection{Data Collection from OpenStreetMap}

We acquire bridge data exclusively from OpenStreetMap via the Overpass API~\cite{overpass}, 
using two complementary query patterns:

\begin{enumerate}
    \item \texttt{\{"man\_made": "bridge"\}}: Standalone bridge structures
    \item \texttt{\{"bridge": True\}}: Road/railway segments with bridge tag
\end{enumerate}

\textbf{Filtering Strategy:} To focus on municipally-managed infrastructure relevant 
for policy decisions, we filter to \emph{named bridges only} (OSM \texttt{name} 
attribute present and non-empty). This eliminates approximately 75\% of unnamed 
minor crossings (pedestrian bridges, small culverts) while retaining all major 
infrastructure. Named railway bridges are included, as many overpasses are 
municipally maintained (e.g., Keio Line Overpass).

\textbf{Comparison with Prior Approaches:} Previous studies relied on municipal 
Excel-based bridge inventories, limiting reproducibility. Our OSM-only 
approach achieved 92.6\% successful bridge-to-road network snapping in Tama City 
(327/353 bridges), validating data quality sufficiency.

\textbf{Additional Geospatial Data:} We retrieve complementary datasets via OSMnx:
\begin{itemize}
    \item \textbf{Road networks}: Drive-accessible street graphs with topological connectivity
    \item \textbf{Buildings}: Residential and commercial structures for impact assessment
    \item \textbf{Public facilities}: Hospitals (OSM \texttt{amenity=hospital}), 
    bus stops (\texttt{highway=bus\_stop}), parks (\texttt{leisure=park}), shops 
    (\texttt{shop=*})
    \item \textbf{Elevation data}: Digital Elevation Models (DEM) for rural area 
    identification (elevation $\geq 100$m threshold)
\end{itemize}

\subsubsection{Heterogeneous Graph Construction}

We construct a heterogeneous graph $G = (V, E)$ with multiple node and edge types:

\textbf{Node Types:}
\begin{itemize}
    \item $V_{\text{bridge}}$: Bridge nodes (353 in Tama City, subset in Morioka)
    \item $V_{\text{street}}$: Road network intersection nodes (typical: 60,000--120,000 nodes)
    \item $V_{\text{building}}$: Residential/commercial building centroids
    \item $V_{\text{hospital}}$: Medical facility locations
    \item $V_{\text{bus\_stop}}$: Public transit access points
    \item $V_{\text{park}}$: Green space/environmental amenities
    \item $V_{\text{shop}}$: Commercial establishments for supply chain analysis
\end{itemize}

\textbf{Edge Types:}
\begin{itemize}
    \item $E_{\text{street-street}}$: Road network connectivity (from OSM topology)
    \item $E_{\text{bridge-street}}$: Bridge-to-road connections via $k$-nearest 
    neighbor ($k=3$, max distance 30m). Each bridge connects to its 3 closest street 
    nodes, enabling robust connectivity even when bridge centroids are slightly 
    offset from road network nodes.
    \item $E_{\text{street-bridge}}$: Bidirectional reverse edges for message passing
    \item $E_{\text{bridge-building}}$: Proximity-based edges (within 1000m radius)
    \item $E_{\text{bridge-facility}}$: Proximity to hospitals, bus stops, parks (varied radii)
\end{itemize}

\textbf{Design Rationale:} The $k$-NN approach ($k=3$) provides more robust topology 
than direct graph snapping, which fails when bridges are offset from road network 
nodes. Bidirectional edges enable information flow between street network structure 
and bridge nodes, critical for centrality-based features.

\subsubsection{Social Impact Indicators}

We compute five complementary social impact indicators, each capturing distinct 
urban functions. All indicators are normalized to [0, 100] scale for comparability.

\paragraph{Transit Desert Score}

\textbf{Definition:} Quantifies degradation of public transit accessibility when 
a bridge $b$ closes, measured by the number of residences losing access to nearest 
bus stops.

\textbf{Mathematical Formulation:}

Let $R_b = \{r_1, r_2, \ldots, r_n\}$ be the set of residences within radius $d_{\text{impact}} = 5$\,km of bridge $b$, and $B = \{b_1, b_2, \ldots, b_m\}$ be the set of bus stops. For each residence $r_i$, define:

\begin{equation}
\begin{split}
d_{\text{pre}}(r_i) &= \min_{j \in [1,k]} \text{dist}(r_i, b_j) \\
&\quad \text{(shortest path to $k=5$ nearest bus stops)}
\end{split}
\end{equation}

Under bridge closure simulation (edge set $E \to E \setminus E_b$ where $E_b$ are edges incident to $b$):

\begin{equation}
d_{\text{post}}(r_i) = \min_{j \in [1,k]} \text{dist}_{G \setminus b}(r_i, b_j)
\end{equation}

A residence is \emph{affected} if path increase exceeds threshold $\theta = 500$\,m:

\begin{equation}
\mathbb{1}_{\text{affected}}(r_i) = \begin{cases}
1 & \text{if } d_{\text{post}}(r_i) - d_{\text{pre}}(r_i) > \theta \\
0 & \text{otherwise}
\end{cases}
\end{equation}

The transit desert score is:

\begin{equation}
S_{\text{transit}}(b) = \frac{\sum_{i=1}^{n} \mathbb{1}_{\text{affected}}(r_i)}{N_{\text{norm}}} \times 100
\end{equation}

where $N_{\text{norm}} = 500$ is the normalization constant.

\textbf{Computation:}
\begin{enumerate}
    \item Sample up to 300 residential buildings within 5km radius using KDTree ($O(\log n)$ query)
    \item For each residence, compute shortest path distance to $k=5$ nearest bus stops
    \item Simulate bridge closure by removing bridge node and incident edges
    \item Recompute shortest paths; count residences with path length increase $>$ 500m threshold
    \item Apply normalization formula above
\end{enumerate}

\textbf{Optimization:} KDTree spatial indexing with 5km impact 
radius achieves 99.5\% computation reduction vs.\ naive all-pairs approach 
(118 hours $\to$ 7.6 minutes).

\paragraph{Hospital Access Score}

\textbf{Definition:} Impact on medical facility accessibility, measured by degradation 
of emergency/routine healthcare access for residential populations.

\textbf{Mathematical Formulation:}

Let $H = \{h_1, h_2, \ldots, h_p\}$ be the set of hospitals. For each residence $r_i$, define the weighted hospital access degradation:

\begin{equation}
d_{\text{hosp}}(r_i) = \sum_{j=1}^{k} w_j \cdot \left( \text{dist}_{G \setminus b}(r_i, h_j) - \text{dist}_G(r_i, h_j) \right)
\end{equation}

where $w_j = \frac{1}{j}$ is the inverse-rank weight (nearest hospital weighted highest), $k=3$.

The hospital access score aggregates across all affected residences:

\begin{equation}
S_{\text{hospital}}(b) = \frac{\sum_{r_i \in R_b'} d_{\text{hosp}}(r_i)}{N_{\text{res}} \cdot D_{\text{norm}}} \times 100
\end{equation}

where $R_b'$ is the KDTree-filtered set of residences within spatial influence of bridge $b$, $N_{\text{res}}$ is the total residential count, and $D_{\text{norm}}$ is a normalization distance constant.

\textbf{Computation:}
\begin{enumerate}
    \item Identify all residential buildings (typical: 3,276 in mid-sized city)
    \item For each residence, find $k=3$ nearest hospitals
    \item Apply KDTree pre-filtering: only consider residences within spatial 
    distance threshold of bridge, achieving 93\% computation reduction
    \item Simulate closure and measure weighted path length increase using equation above
    \item Apply normalization to [0, 100] scale
\end{enumerate}

\textbf{Optimization:} KDTree pre-filtering retains only residences within spatial 
influence of each bridge, achieving 93\% computation reduction 
(13 hours $\to$ 8.7 minutes).

\paragraph{Isolation Risk Score}

\textbf{Definition:} Risk of community isolation in rural/mountainous areas with 
limited alternative routes, measured by alternative path availability for high-elevation 
populations.

\textbf{Mathematical Formulation:}

Let $R_{\text{rural}} = \{r \in R : \text{elevation}(r) \geq \theta_{\text{elev}}\}$ be rural residences (threshold $\theta_{\text{elev}} = 100$\,m). After bridge $b$ closure, perform connected component decomposition:

\begin{equation}
G \setminus b = \bigcup_{i=1}^{C} G_i \quad \text{(disjoint connected components)}
\end{equation}

A rural residence $r_j$ is \emph{isolated} if it becomes disconnected from urban core nodes $U$:

\begin{equation}
\begin{split}
\mathbb{1}_{\text{isolated}}(r_j) &= \begin{cases}
1 & \text{if } \nexists \text{ path from } r_j \text{ to any } \\
  & \quad u \in U \text{ in } G \setminus b \\
0 & \text{otherwise}
\end{cases}
\end{split}
\end{equation}

The isolation risk score is:

\begin{equation}
S_{\text{isolation}}(b) = \frac{\sum_{r_j \in R_{\text{rural}}} \mathbb{1}_{\text{isolated}}(r_j) \cdot \text{pop}(r_j)}{P_{\text{total}}} \times 100
\end{equation}

where $\text{pop}(r_j)$ is estimated population at residence $r_j$ and $P_{\text{total}}$ is total population.

\textbf{Computation:}
\begin{enumerate}
    \item Identify rural areas: DEM elevation $\geq 100$m threshold (configurable 
    per region)
    \item For each bridge, detect nearby rural residential clusters within 3km radius
    \item Simulate closure and perform connected component analysis using DFS/BFS
    \item Apply indicator function to count isolated residences
    \item Weight by population and normalize to [0, 100]
\end{enumerate}

\textbf{Optimization:} KDTree 3km impact radius and elevation-based pre-filtering. 
Execution time: 5--10 minutes.

\paragraph{Supply Chain Impact Score}

\textbf{Definition:} Economic disruption to commercial logistics via degraded access 
to primary/trunk highways (national road network).

\textbf{Mathematical Formulation:}

Let $S = \{s_1, s_2, \ldots, s_q\}$ be commercial establishments and $H_{\text{hwy}} = \{h_1, h_2, \ldots, h_m\}$ be highway nodes (OSM \texttt{highway=primary|trunk}). For each shop $s_i$:

\begin{equation}
\begin{split}
d_{\text{supply}}(s_i) &= \sum_{j=1}^{k} w_{\text{cat}}(s_i) \cdot \\
&\quad \left( \text{dist}_{G \setminus b}(s_i, h_j) - \text{dist}_G(s_i, h_j) \right)
\end{split}
\end{equation}

where $k=3$ nearest highways, and $w_{\text{cat}}(s_i)$ is category weight:

\begin{equation}
\begin{split}
w_{\text{cat}}(s_i) &= \begin{cases}
1.5 & \text{if shop category} \in \\
    & \quad \{\text{food}, \text{daily necessities}\} \\
1.0 & \text{otherwise}
\end{cases}
\end{split}
\end{equation}

The supply chain impact score is:

\begin{equation}
S_{\text{supply}}(b) = \frac{\sum_{s_i \in S_b} d_{\text{supply}}(s_i)}{N_{\text{shops}} \cdot D_{\text{norm}}} \times 100
\end{equation}

where $S_b$ is shops within influence radius of bridge $b$.

\textbf{Computation:}
\begin{enumerate}
    \item Identify commercial establishments (OSM \texttt{shop=*} tag, typical: 
    194 shops)
    \item For each shop, compute shortest path to $k=3$ nearest highway nodes 
    using Dijkstra's algorithm
    \item Simulate bridge closure and measure path length increase
    \item Apply category weights (food/daily necessities weighted 1.5×)
    \item Normalize to [0, 100] using formula above
\end{enumerate}

\textbf{Optimization:} Highway node pre-identification and spatial indexing. 
Execution time: 9.6 minutes.

\paragraph{Green Space Access Score}

\textbf{Definition:} Environmental connectivity and quality-of-life impact, measured 
by degraded access to parks, nature reserves, and recreational areas.

\textbf{Mathematical Formulation:}

Let $P = \{p_1, p_2, \ldots, p_v\}$ be parks/green spaces. For each residence $r_i$ in large residential set $R$ (typical: $|R| = 10{,}372$):

\begin{equation}
\begin{split}
d_{\text{green}}(r_i) &= \sum_{j=1}^{k} \frac{1}{j} \cdot \\
&\quad \max\left(0, \text{dist}_{G \setminus b}(r_i, p_j) - \text{dist}_G(r_i, p_j)\right)
\end{split}
\end{equation}

where $k=3$ nearest parks with inverse-rank weighting. The green space score aggregates across all residences:

\begin{equation}
S_{\text{green}}(b) = \frac{\sum_{r_i \in R} d_{\text{green}}(r_i)}{|R| \cdot D_{\text{norm}}} \times 100
\end{equation}

\textbf{Computational Note:} Large $|R|$ causes this indicator to dominate pipeline time (43.5 min, 32\% of total). Residential mesh aggregation could achieve 50--100$\times$ additional speedup:

\begin{equation}
\begin{split}
R &\to R_{\text{mesh}} = \{m_1, m_2, \ldots, m_M\} \\
&\quad \text{where } |R_{\text{mesh}}| \ll |R|
\end{split}
\end{equation}

with population-weighted centroids could achieve 50--100$\times$ additional speedup.

\textbf{Computation:}
\begin{enumerate}
    \item Identify all parks/green spaces (OSM \texttt{leisure=park}, 
    \texttt{leisure=nature\_reserve})
    \item For all $|R| = 10{,}372$ residences, compute weighted paths 
    to $k=3$ nearest parks using formula above
    \item Simulate closure and measure path increase
    \item Normalize to [0, 100]
\end{enumerate}

\paragraph{Composite Social Impact Score}

The overall bridge importance is computed as a weighted combination of all five indicators:

\begin{equation}
S_{\text{overall}}(b) = \sum_{i=1}^{5} \alpha_i S_i(b)
\end{equation}

where $S_i \in \{S_{\text{transit}}, S_{\text{hospital}}, S_{\text{isolation}}, S_{\text{supply}}, S_{\text{green}}\}$

and default weights are $\alpha_i = 0.2$ (equal weighting):

\begin{equation}
\begin{split}
\boldsymbol{\alpha} &= [\alpha_{\text{transit}}, \alpha_{\text{hospital}}, \alpha_{\text{isolation}}, \\
&\quad \alpha_{\text{supply}}, \alpha_{\text{green}}] \\
&= [0.2, 0.2, 0.2, 0.2, 0.2]
\end{split}
\end{equation}

\textbf{Configurable Weighting:} Alternative schemes can be configured for city-specific priorities:

\begin{itemize}
    \item \textbf{Aging population emphasis}: Increase $\alpha_{\text{hospital}} = 0.4$, reduce others to 0.15
    \item \textbf{Rural isolation focus}: Increase $\alpha_{\text{isolation}} = 0.35$, $\alpha_{\text{transit}} = 0.25$
    \item \textbf{Economic priority}: Increase $\alpha_{\text{supply}} = 0.35$, maintain healthcare baseline
\end{itemize}

subject to normalization constraint $\sum_{i=1}^{5} \alpha_i = 1$.

\textbf{Important Design Note:} We do \emph{not} apply log1p transformation to 
these bounded scores $S_i \in [0, 100]$. Prior experiments (v1.6 HGNN studies) demonstrated 
that log-transforms degrade performance for bounded metrics with modest variance:

\begin{equation}
\text{log1p transform} \quad \tilde{S}_i = \log(1 + S_i) \quad \textbf{NOT USED}
\end{equation}

Unlike unbounded network centrality measures (betweenness: $[0, \infty)$ with heavy right skew) where log-transforms handle long-tail distributions, our bounded social impact scores exhibit moderate skewness suitable for direct use.

\subsubsection{Computational Optimization}

Table~\ref{tab:optimization} summarizes computational improvements from naive 
baseline (v1.5.0) to optimized version (v1.5.3):

\begin{table*}[ht]
\centering
\caption{Computational Optimization Results}
\label{tab:optimization}
\small
\begin{tabular}{lccc}
\toprule
\textbf{Phase} & \textbf{v1.5.0 (naive)} & \textbf{v1.5.3 (optimized)} & \textbf{Speedup} \\
\midrule
Transit Desert & 118 hours (est.) & 7.6 min & 933$\times$ \\
Hospital Access & 13 hours (est.) & 8.7 min & 90$\times$ \\
Isolation Risk & 2 hours (est.) & 5--10 min & 12--24$\times$ \\
Supply Chain & -- & 9.6 min & N/A \\
Green Space & -- & 43.5 min & N/A \\
\midrule
\textbf{Total} & \textbf{133 hours} & \textbf{20--25 min} & \textbf{~40$\times$} \\
\bottomrule
\end{tabular}
\end{table*}

Key optimization strategies:
\begin{itemize}
    \item \textbf{KDTree Spatial Indexing:} Replace all-pairs distance computations 
    $O(n^2)$ with logarithmic-time nearest-neighbor queries $O(\log n)$. For point set $P = \{p_1, p_2, \ldots, p_n\}$, KDTree partitions space via recursive median splits:
    
    \begin{equation}
    \text{KDTree}(P, d) = \begin{cases}
    \text{leaf}(P) & |P| \leq \ell \\
    \text{node}(...) & \text{otherwise}
    \end{cases}
    \end{equation}
    
    where split dimension cycles through coordinates, $P_L$ and $P_R$ are left/right partitions by median.
    
    \textbf{Query complexity:} $k$-NN search in KDTree: $O(k \log n)$ vs. naive $O(kn)$
    
    \item \textbf{Impact Radius Constraints:} Limit analysis to geographically 
    relevant areas. For bridge $b$ at position $\mathbf{x}_b$, define influence set:
    
    \begin{equation}
    R_b^{(r)} = \{r_i \in R : \|\mathbf{x}_{r_i} - \mathbf{x}_b\|_2 \leq d_{\text{max}}\}
    \end{equation}
    
    where $d_{\text{max}} \in \{5\text{km (transit)}, 3\text{km (isolation)}\}$ prunes 95--99\% of residences.
    
    \item \textbf{Pre-filtering:} Eliminate unaffected entities before expensive 
    path calculations using spatial distance bounds
    \item \textbf{Graph Edge Manipulation:} Simulate closure by setting edge weights to $\infty$ rather than creating graph copies, reducing memory overhead from $O(|V| \cdot |E|)$ to $O(|E_b|)$ where $|E_b|$ is edges incident to bridge $b$
\end{itemize}

Result: 40$\times$ overall speedup (133 hours $\to$ 20--25 minutes for full city analysis).

\subsection{Functional Typology Discovery via Clustering}
\label{subsec:clustering}

\subsubsection{Feature Engineering}

Each bridge is represented by a 52-dimensional feature vector combining:

\textbf{Social Impact Scores (5 dimensions):}
\begin{itemize}
    \item Transit desert score, hospital access score, isolation risk score, 
    supply chain impact score, green space access score
\end{itemize}

\textbf{Network Topology Features (8 dimensions):}
\begin{itemize}
    \item Betweenness centrality, num\_street\_connections, 2-hop neighbor count, 
    clustering coefficient, metapath features (street degree average, etc.)
\end{itemize}

\textbf{Spatial Features (6 dimensions):}
\begin{itemize}
    \item Distance to river/coast (log-scale), elevation, latitude, longitude
\end{itemize}

\textbf{Facility Proximity Features (15 dimensions):}
\begin{itemize}
    \item Counts/distances to hospitals, schools, bus stops, parks, shops within 
    various radii
\end{itemize}

\textbf{Bridge Attributes (18 dimensions):}
\begin{itemize}
    \item OSM tags (highway type, bridge type, railway crossing), encoded as 
    one-hot or numeric features
\end{itemize}

\textbf{Feature Vector Representation:}

Each bridge $b$ is represented by feature vector $\mathbf{x}_b \in \mathbb{R}^{52}$:

\begin{equation}
\begin{split}
\mathbf{x}_b = [&\underbrace{S_{\text{transit}}, S_{\text{hospital}}, \ldots, S_{\text{green}}}_{\text{5 social impact}}, \\
&\underbrace{\text{betweenness}, \text{degree}, \ldots}_{\text{8 topology}}, \\
&\underbrace{\text{lat}, \text{lon}, \text{elev}, \ldots}_{\text{6 spatial}}, \ldots]^T
\end{split}
\end{equation}

\textbf{Preprocessing Pipeline:}

\begin{enumerate}
    \item \textbf{Zero-variance filtering:} Remove features with $\sigma^2 = 0$ (23 features in Tama City):
    \begin{equation}
    \mathcal{F}_{\text{retained}} = \{f_i : \text{Var}(f_i) > \epsilon\} \quad \text{where } \epsilon = 10^{-6}
    \end{equation}
    
    \item \textbf{Z-score normalization:} For each feature $f_i$, compute mean $\mu_i$ and std. dev. $\sigma_i$ across all bridges, then normalize:
    \begin{equation}
    \begin{split}
    \tilde{x}_{bi} &= \frac{x_{bi} - \mu_i}{\sigma_i} \quad \text{where} \\
    \mu_i &= \frac{1}{N}\sum_{j=1}^{N} x_{ji}, \\
    \sigma_i &= \sqrt{\frac{1}{N}\sum_{j=1}^{N} (x_{ji} - \mu_i)^2}
    \end{split}
    \end{equation}
    
    This ensures all features contribute equally to distance metrics: $\tilde{x}_{bi} \sim \mathcal{N}(0, 1)$ approximately.
    
    \item \textbf{Outlier detection:} Flag bridges with extreme feature values. Define outlier indicator:
    \begin{equation}
    \mathbb{1}_{\text{outlier}}(b) = \begin{cases}
    1 & \text{if } \sum_{i=1}^{52} \mathbb{1}[|\tilde{x}_{bi}| > 3] > 3 \\
    0 & \text{otherwise}
    \end{cases}
    \end{equation}
    
    12 outliers detected (1.55\% in 775-bridge dataset) $\to$ flagged but retained in analysis
\end{enumerate}

\subsubsection{UMAP Dimensionality Reduction}

UMAP~\cite{mcinnes2018umap} projects the 52-dimensional feature space into 2D 
for visualization while preserving both local neighborhoods and global manifold structure.

\textbf{Theoretical Foundation:}

UMAP constructs a fuzzy topological representation of high-dimensional data. For each point $x_i$, define local connectivity via:

\begin{equation}
w(x_i, x_j) = \exp\left(-\frac{\max(0, d(x_i, x_j) - \rho_i)}{\sigma_i}\right)
\end{equation}

where $\rho_i$ is distance to nearest neighbor, $\sigma_i$ is normalization parameter ensuring fixed number of neighbors (\texttt{n\_neighbors} = 15).

The high-dimensional fuzzy set representation $A$ and low-dimensional embedding $B$ are optimized via cross-entropy:

\begin{equation}
C = \sum_{i,j} w_{ij}^{(A)} \log\frac{w_{ij}^{(A)}}{w_{ij}^{(B)}} + (1-w_{ij}^{(A)}) \log\frac{1-w_{ij}^{(A)}}{1-w_{ij}^{(B)}}
\end{equation}

Optimized via stochastic gradient descent with negative sampling.

\textbf{Hyperparameters:}
\begin{itemize}
    \item \textbf{n\_neighbors = 15}: Number of neighbors for local structure 
    preservation. Controls local vs. global emphasis (larger values $\to$ more 
    global structure). Sensitivity: \textbf{High}. Tested range: [10, 15, 20]; 
    15 provides balanced representation.
    \item \textbf{min\_dist = 0.1}: Minimum distance between points in embedding 
    space $d_{\text{min}}$. Controls tightness of clusters: smaller values $\to$ tighter clusters.
    \item \textbf{metric = euclidean}: Distance metric $d(x_i, x_j) = \|x_i - x_j\|_2$ in feature space (post-normalization)
    \item \textbf{n\_components = 2}: Output dimensionality for visualization
\end{itemize}

\textbf{Why UMAP over t-SNE?}
UMAP preserves both local and global manifold topology (unlike t-SNE), runs
in $O(N \log N)$ time vs.\ $O(N^2)$, and has theoretical grounding in
Riemannian geometry. These properties favor 52D $\to$ 2D reduction with
775 samples.

\subsubsection{HDBSCAN Clustering}

HDBSCAN~\cite{mcinnes2017hdbscan} performs density-based clustering on the 2D 
UMAP embedding, identifying functional bridge archetypes.

\textbf{Algorithmic Foundation:}

HDBSCAN extends DBSCAN by building a cluster hierarchy. For points $\{y_1, y_2, \ldots, y_n\}$ in 2D UMAP space:

\begin{enumerate}
    \item Compute mutual reachability distance:
    \begin{equation}
    \begin{split}
    d_{\text{mreach}}(y_i, y_j) &= \max\{\text{core}_k(y_i), \\
    &\quad \text{core}_k(y_j), d(y_i, y_j)\}
    \end{split}
    \end{equation}
    where $\text{core}_k(y_i)$ is distance to $k$-th nearest neighbor (\texttt{min\_samples} = 10).
    
    \item Build minimum spanning tree (MST) on mutual reachability graph
    
    \item Convert MST to cluster hierarchy by cutting edges in ascending $d_{\text{mreach}}$ order
    
    \item Extract stable clusters using Excess of Mass (EOM) method:
    \begin{equation}
    \text{stability}(C) = \sum_{y \in C} (\lambda_{\text{death}}(C) - \lambda_{\text{birth}}(y))
    \end{equation}
    where $\lambda = 1/d_{\text{mreach}}$ is inverse distance, measuring cluster persistence across density levels.
\end{enumerate}

Clusters with \texttt{stability} above threshold and size $\geq$ \texttt{min\_cluster\_size} = 20 are extracted. Points not in any cluster receive noise label $-1$.

\textbf{Hyperparameters:}
\begin{itemize}
    \item \textbf{min\_cluster\_size = 20}: Minimum bridges forming a cluster. 
    Tested range: [10, 20, 30]; 20 balances granularity with stability.
    \item \textbf{min\_samples = 10}: Core point definition in mutual reachability distance
    \item \textbf{cluster\_selection\_method = `eom'}: Excess of Mass extraction from hierarchy
\end{itemize}

\textbf{Why HDBSCAN over k-means?}
HDBSCAN requires no pre-specified cluster count, handles arbitrary cluster shapes
and varying densities, and explicitly identifies outliers (noise label $= -1$)
rather than forcing every point into a cluster.
Detailed clustering results are reported in Section~\ref{sec:results}.

\subsubsection{Visualization and Statistical Profiling}

\begin{figure}[t]
\centering
\includegraphics[width=0.48\textwidth]{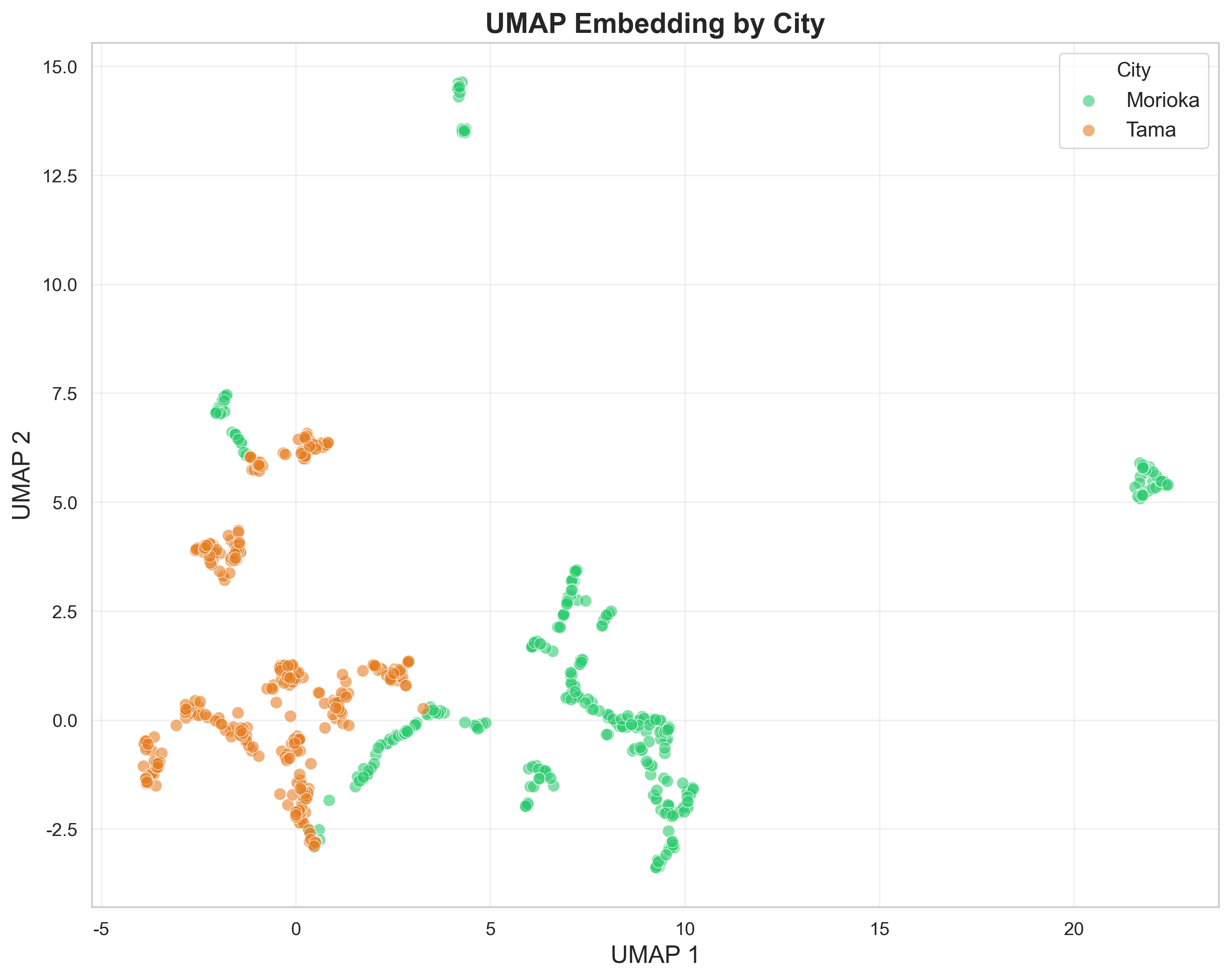}
\caption{UMAP visualization colored by city (Tama City vs. Morioka City). 
Near-perfect geographic separation demonstrates that urban context (dense 
metropolitan vs. regional mixed terrain) is the primary determinant of bridge 
functional profiles, validating context-dependency hypothesis.}
\label{fig:umap_city}
\end{figure}

\begin{figure}[t]
\centering
\includegraphics[width=0.48\textwidth]{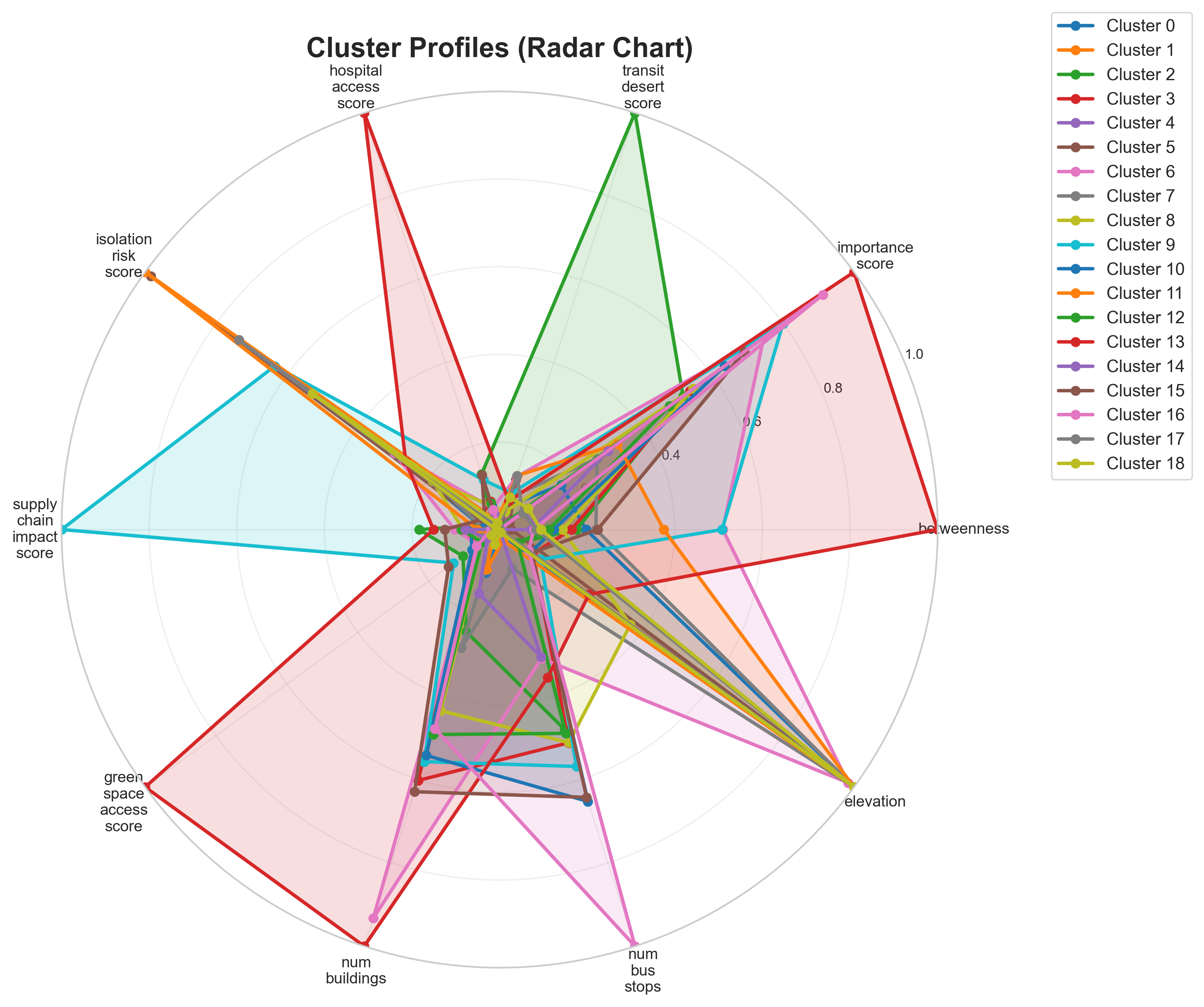}
\caption{Radar charts comparing normalized z-scores for four representative clusters. 
Each axis represents one of the five social impact indicators. Cluster-specific 
shapes enable targeted maintenance strategies.}
\label{fig:radar_profiles}
\end{figure}

\begin{figure*}[t]
\centering
\includegraphics[width=0.32\textwidth]{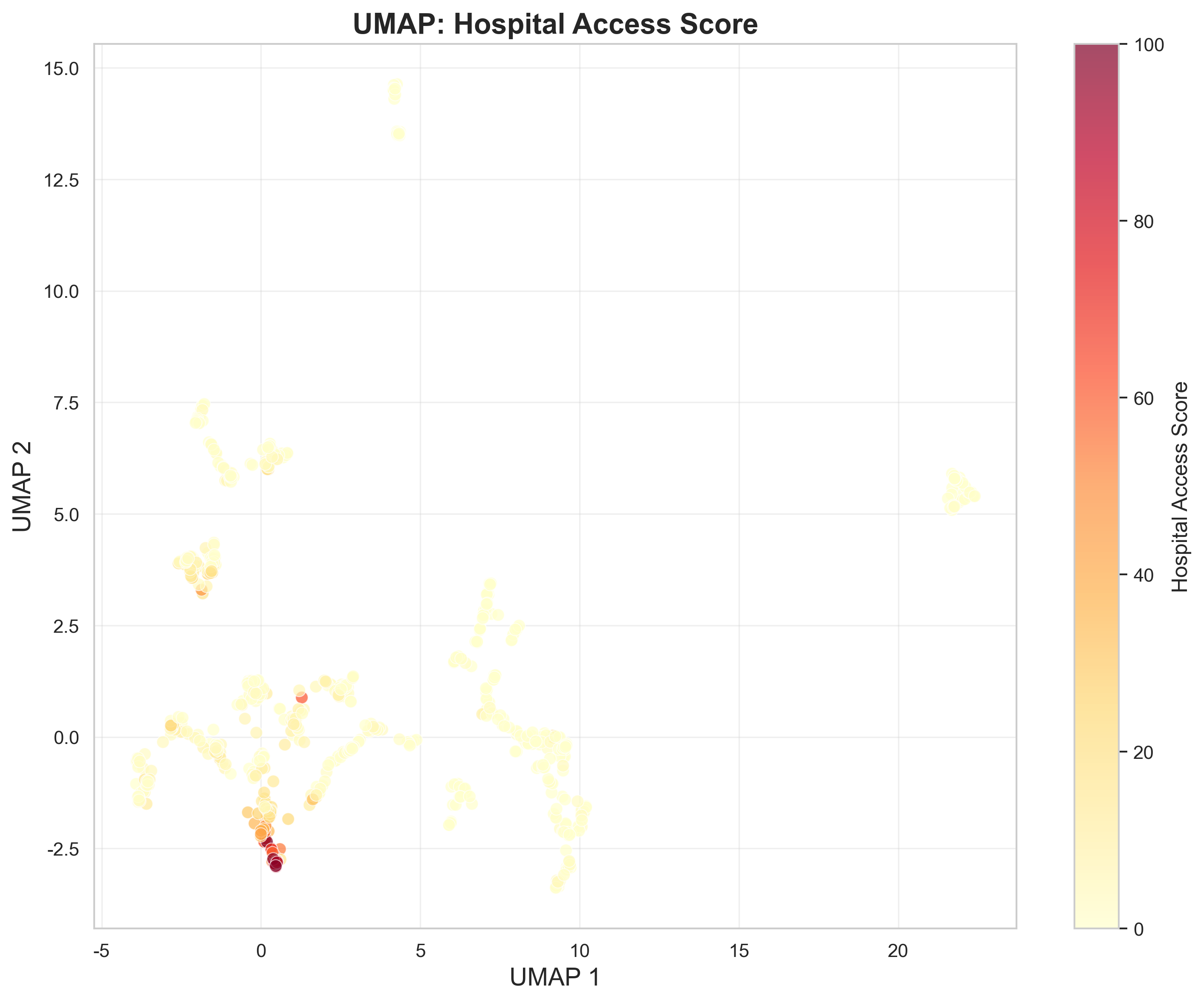}
\hfill
\includegraphics[width=0.32\textwidth]{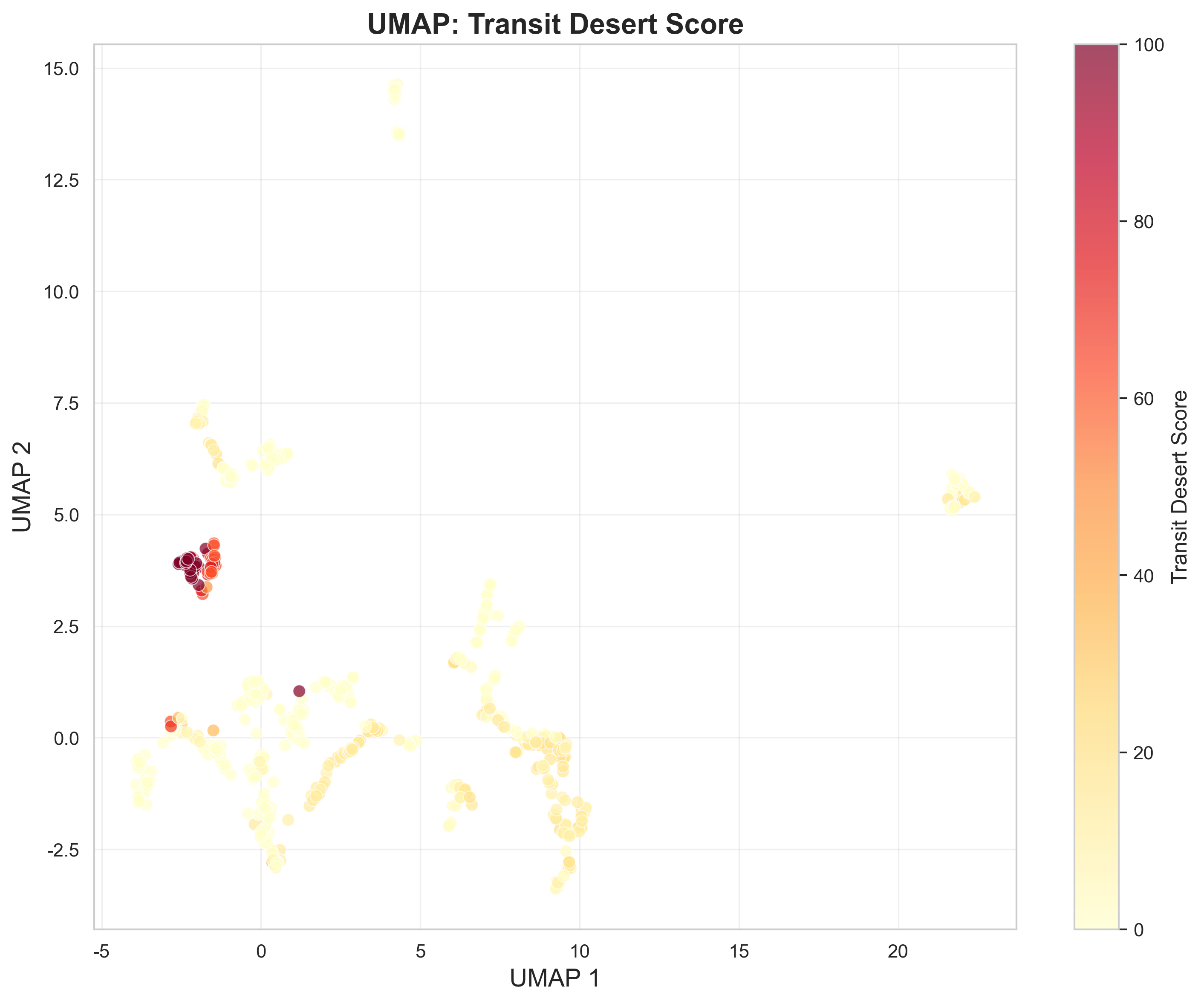}
\hfill
\includegraphics[width=0.32\textwidth]{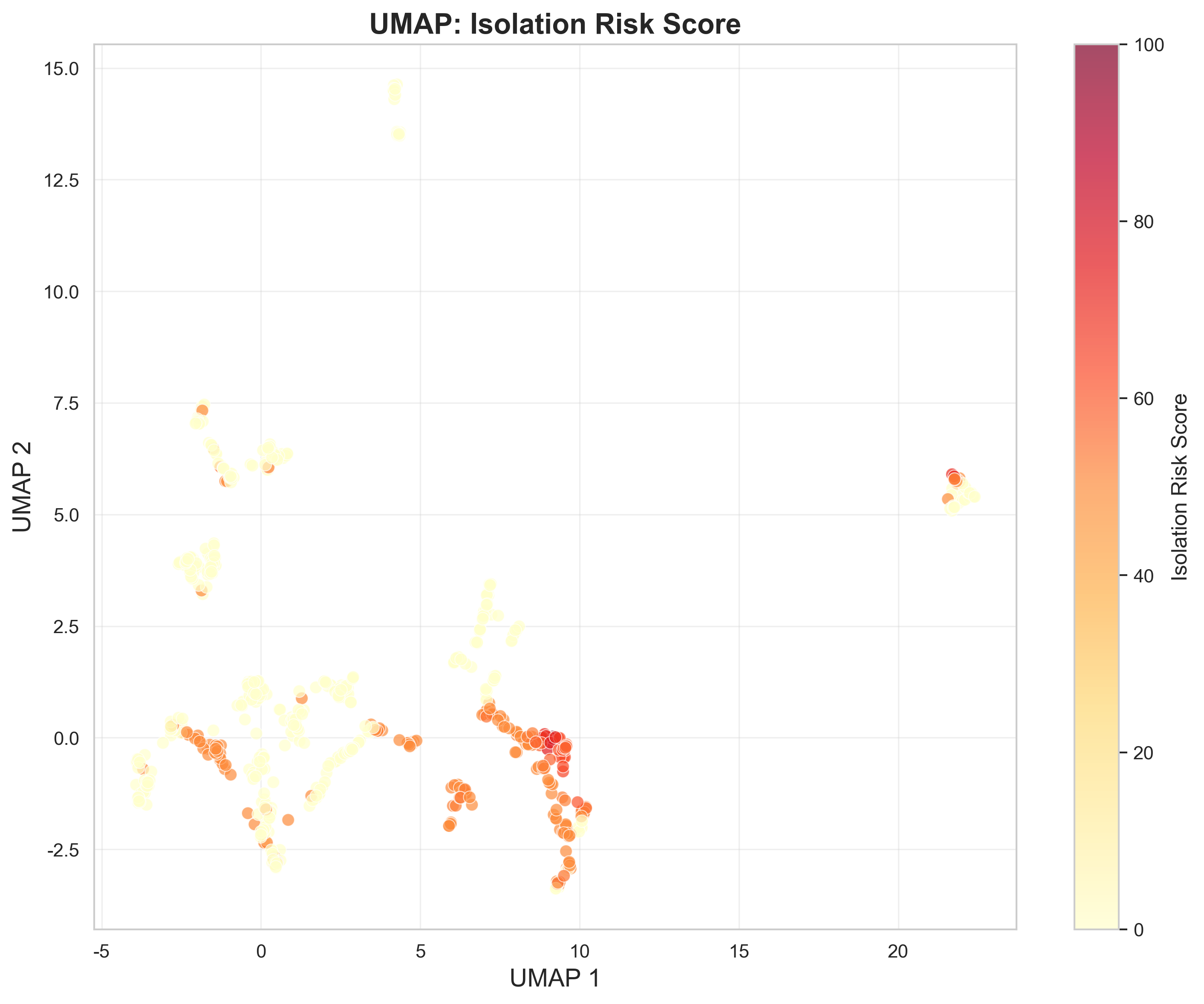}
\caption{UMAP heatmaps overlaying social impact scores on embedding space: 
(left) hospital access, (center) transit desert, (right) isolation risk. 
Warmer colors = higher scores. Spatial gradients validate that UMAP preserves 
indicator-specific structure in dimensionality reduction.}
\label{fig:umap_heatmaps}
\end{figure*}

We generate four visualization types:

\begin{enumerate}
    \item \textbf{UMAP Scatter Plot (by cluster):} 2D projection with 19 cluster 
    colors, enabling visual inspection of cluster separation and density patterns 
    (Figure~\ref{fig:umap_cluster})
    
    \item \textbf{UMAP Scatter Plot (by city):} Same projection colored by Tama/Morioka, 
    revealing geographic clustering pattern (Figure~\ref{fig:umap_city})
    
    \item \textbf{Feature Heatmap:} 19 clusters $\times$ 52 features matrix with 
    z-score coloring, identifying characteristic features per cluster 
    (Figure~\ref{fig:umap_heatmaps})
    
    \item \textbf{Radar Charts:} Multidimensional profiles for representative 
    clusters, emphasizing top-3 distinguishing features (Figure~\ref{fig:radar_profiles})
\end{enumerate}

Statistical profiles (mean, std per feature per cluster) are exported to 
\texttt{cluster\_statistics.csv} for downstream LLM interpretation.

\subsection{Automated Interpretation via Temperature-Optimized LLMs}
\label{subsec:llm}

\subsubsection{LLM Model Selection}

We compare two Japanese LLMs fine-tuned on a construction domain corpus 
(kasensabo\_jp dataset, $n{=}4{,}000$ river and sediment control technical standards):

\textbf{Candidate Models:}
\begin{enumerate}
    \item \textbf{Swallow-8B-LoRA} (TokyoTech/llm-jp lineage): Emphasizes structural 
    engineering and construction science perspective. Example terminology: 
    ``power supply load'' (denryoku kyoukyu fuka), ``total combined weight'' 
    (sougou kajuu).
    
    \item \textbf{Elyza-8B-LoRA} (ELYZA Inc.): Emphasizes urban operations and 
    municipal management perspective. Example terminology: ``transport planning'' 
    (koutsuu keikaku), ``disaster safety function'' (saigaiji anzen kinou).
\end{enumerate}

\textbf{Evaluation Criteria:} Qualitative assessment across 19 clusters on:
\begin{itemize}
    \item \textbf{Urban function perspective} (High weight): Alignment with policy/planning 
    needs vs. structural engineering focus
    \item \textbf{Policy/civic appropriateness} (High weight): Suitability for 
    stakeholder communication
    \item \textbf{Quantitative grounding} (Medium weight): Use of statistical evidence 
    from feature data
    \item \textbf{Structural/technical depth} (Low weight): Construction domain expertise
\end{itemize}

\textbf{Selection Result:} \textbf{Elyza-8B was selected} based on superior urban 
function expression (18/19 clusters). While Swallow-8B provides deeper structural 
analysis, Elyza-8B's operational perspective better serves policy 
communication needs.

\subsubsection{Temperature Parameter Optimization}

Temperature $T$ controls sampling randomness in LLM token generation via softmax distribution:

\begin{equation}
P(w_i | \text{context}) = \frac{\exp(\text{logit}_i / T)}{\sum_j \exp(\text{logit}_j / T)}
\end{equation}

where $\text{logit}_i$ is the model's raw output score for token $w_i$. 

Lower $T \to$ sharper distribution (more deterministic, peaks at $\arg\max$), higher $T \to$ flatter distribution (more diverse/random outputs). At limit:

\begin{align}
T \to 0^+: \quad & P(w_i) \to \mathbb{1}[w_i = \arg\max_j \text{logit}_j] \nonumber \\
& \quad \text{(greedy decoding)} \\
T \to \infty: \quad & P(w_i) \to \frac{1}{|V|} \quad \text{(uniform random)}
\end{align}

\textbf{Experimental Design:} Generate interpretations for all 19 clusters at 
four temperature values: $T \in \{0.1, 0.3, 0.5, 0.7\}$. Evaluate on:
\begin{enumerate}
    \item \textbf{Output stability:} Character count variance across 19 clusters
    \item \textbf{Structural consistency:} Presence of all 5 required sections
    \item \textbf{Factual grounding:} Absence of unfounded speculation
    \item \textbf{Error rate:} Typos, grammatical issues
\end{enumerate}

\textbf{Results Summary:}
We find that $T{=}0.3$ yields optimal results: 100\% five-section completeness, 
fact-based reasoning, zero typos, and moderate output length variance (4.0$\times$). 
Surprisingly, $T{=}0.5$ produces the highest instability (6.7$\times$ variance).
$T{=}0.1$ causes over-speculation, while $T{=}0.7$ is stable but verbose.
Detailed comparisons are presented in Section~\ref{sec:results} (Table~\ref{tab:temperature}).

Execution time: 2 minutes 30 seconds for all 19 clusters 
($\approx 6.8$ seconds per cluster).

\subsubsection{Prompt Engineering and Output Structure}

We design a five-section structured template enforcing consistent interpretation format:

\begin{enumerate}
    \item \textbf{Cluster Overview:} Bridge count, city composition, geographic 
    concentration
    
    \item \textbf{Strengths:} Top 3--5 high z-score features with physical 
    and functional interpretation
    
    \item \textbf{Weaknesses:} Low z-score features, associated risks, 
    maintenance recommendations
    
    \item \textbf{Role Classification:} Concise functional archetype 
    label (e.g., Healthcare Access Type)
    
    \item \textbf{City-Specific Context:} Tama vs. Morioka comparison, 
    local policy implications
\end{enumerate}

\textbf{Prompt Template Example:}

\begin{verbatim}
Please interpret the following cluster using
a 5-section structure:

Cluster ID: {cluster_id}
Bridge count: {bridge_count}
City composition: Tama {tama_pct}%, 
                  Morioka {morioka_pct}%

Key features (z-scores):
- hospital_access: {z_score_hospital}
- green_space_access: {z_score_green}
- transit_desert: {z_score_transit}
[...]

Output in the following 5 sections:
1. Cluster Overview
2. Strengths
3. Weaknesses
4. Bridge Role Type
5. City-Specific Context
\end{verbatim}

\subsubsection{Quality Validation}

\textbf{Quantitative Metrics:}
\begin{itemize}
    \item 5-section completeness: 100\% (19/19 clusters, $T=0.3$)
    \item Typo-free rate: 100\% ($T=0.3$)
    \item Mean report length: 290 characters ($T=0.3$)
    \item Total execution time: 2 min 30 sec
\end{itemize}

\textbf{Qualitative Validation:} Manual review confirms cluster role assignments 
align with feature data:
\begin{itemize}
    \item Cluster 2 labeled ``Transit-Connectivity Type'' $\to$ 
    Validated by transit\_desert (z=3.27), isolation\_risk (z=3.14)
    \item Cluster 13 labeled ``Healthcare Access Type'' $\to$ 
    Validated by hospital\_access (z=3.05), green\_space (z=3.03)
    \item Cluster 6 labeled ``Arterial Type'' $\to$ Validated by high 
    passenger lines, building density (z=1.31)
\end{itemize}

\section{Case Studies}
\label{sec:casestudies}

We validate our methodology on two Japanese cities with contrasting characteristics.

\subsection{Tama City Case Study}

\textbf{City Profile:}
\begin{itemize}
    \item Location: Western Tokyo metropolitan area
    \item Area: 99.4 km² (dense urban)
    \item Bridge count: 353 named bridges (224 road + 113 railway + 16 other)
    \item Data source: OpenStreetMap (GSI vector tile service discontinued)
    \item CRS: EPSG:6677 (JGD2011 Plane Rectangular CS Zone 9)
\end{itemize}

\textbf{Data Quality:}
\begin{itemize}
    \item OSM query yield: 922 bridges total $\to$ 353 after named-only filtering 
    (61.7\% reduction)
    \item Bridge-to-street snapping success: 92.6\% (327/353 bridges)
    \item 26 failures: primarily railway bridges and pedestrian bridges not aligned 
    with drivable road network
\end{itemize}

\textbf{Execution Performance:}
\begin{itemize}
    \item Total pipeline time: 135 minutes
    \item Bottleneck: Medical access scoring (61 min, 45\% of total)
    \item Graph construction: 60,529 street nodes, 353 bridge nodes
    \item Cluster distribution: 10 Tama-dominant clusters
\end{itemize}

\textbf{Top-5 Critical Bridges} (social\_impact\_score\_overall):
\begin{enumerate}
    \item Tama Urban Monorail (Tama Toshi Monorail): 77.3
    \item Keio Line Overpass (Keio-sen Koukabashi): 74.7
    \item Keio Line Overpass (Keio-sen Koukabashi): 59.6
    \item Chuo Expressway (Chuo Jidoushado): 46.5
    \item Tama Urban Monorail (Tama Toshi Monorail): 34.8
\end{enumerate}

\textbf{Key Finding:} Railway infrastructure dominates top rankings (4/5 bridges), 
revealing that urban transit hubs are more critical than individual road bridges 
in dense metropolitan areas. This contrasts with rural regions where road bridges 
exhibit higher importance.

\begin{figure}[t]
\centering
\includegraphics[width=0.48\textwidth]{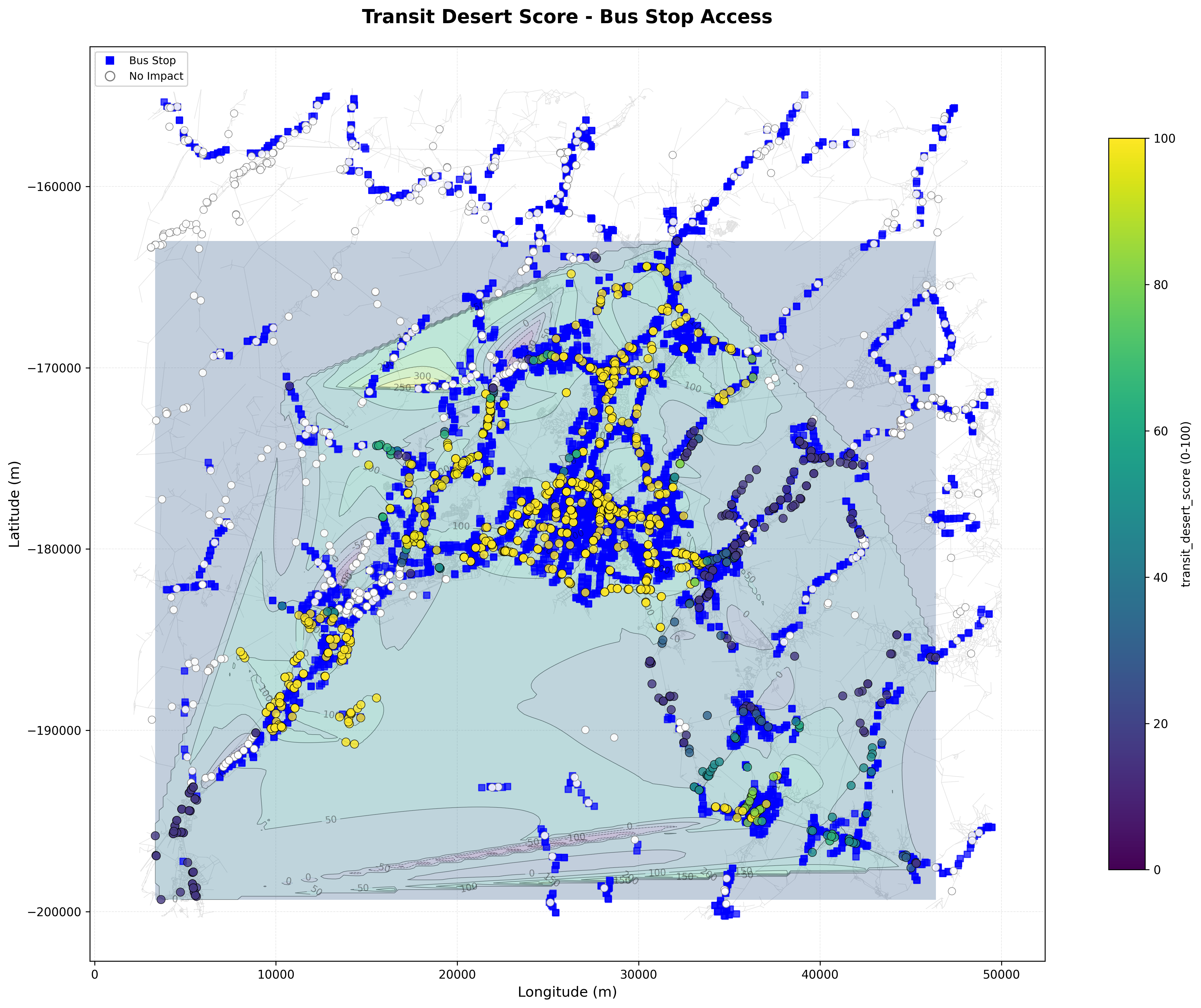}
\caption{Transit desert score map for Tama City showing spatial distribution 
of public transit accessibility. Warmer colors indicate bridges with high transit 
desert scores. Railway overpasses dominate high-score regions.}
\label{fig:tama_transit}
\end{figure}

\begin{figure*}[t]
\centering
\includegraphics[width=0.32\textwidth]{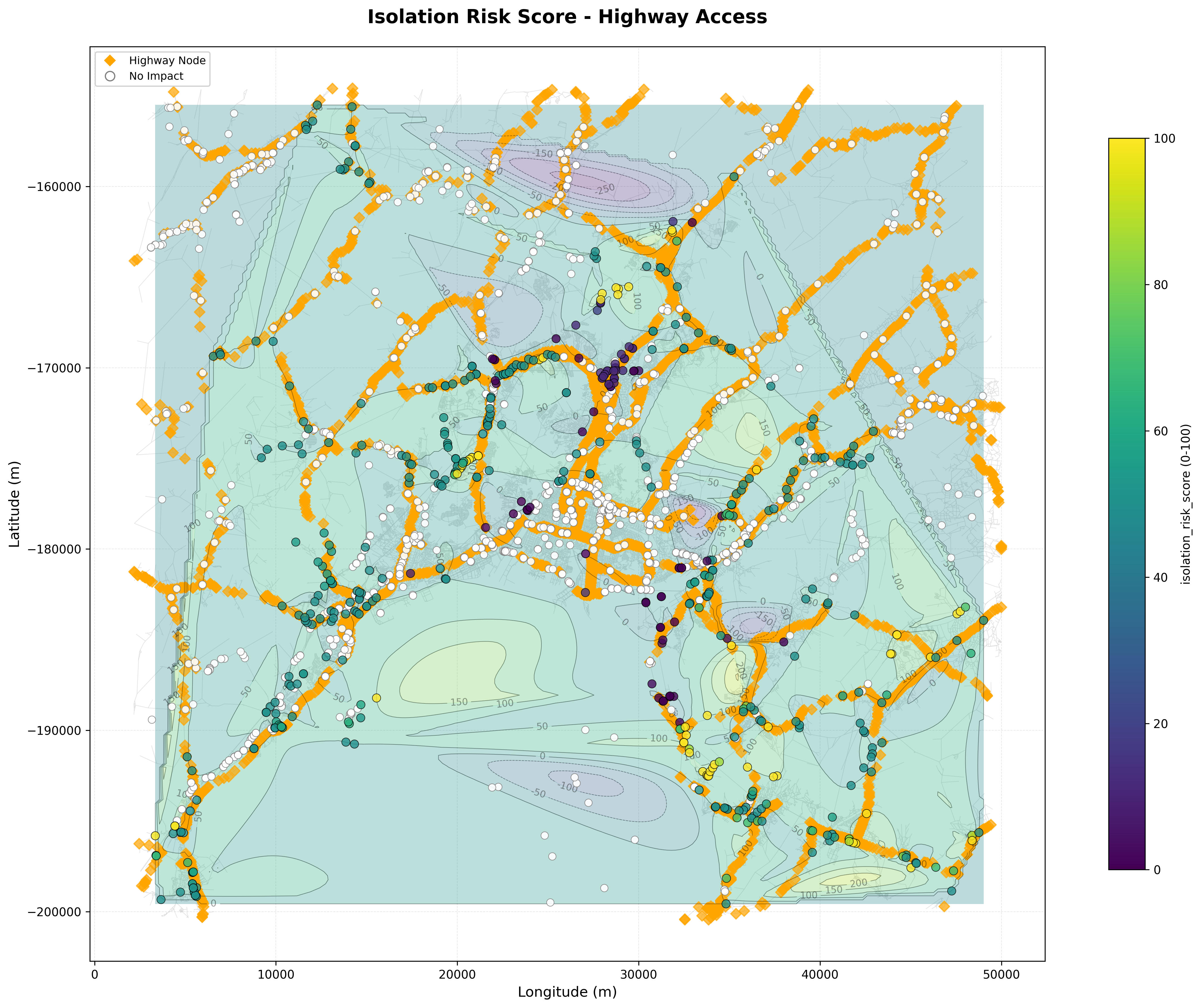}
\hfill
\includegraphics[width=0.32\textwidth]{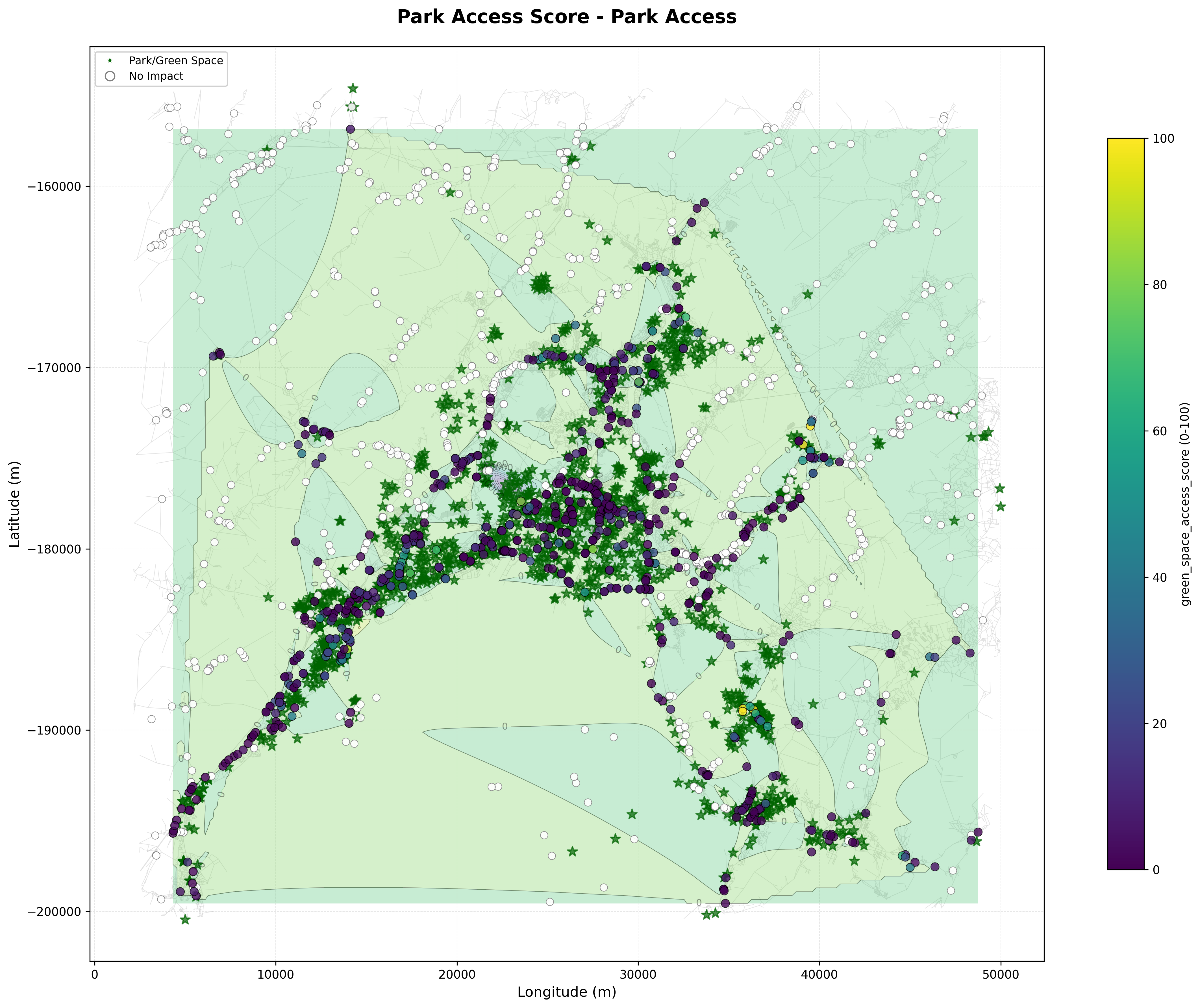}
\hfill
\includegraphics[width=0.32\textwidth]{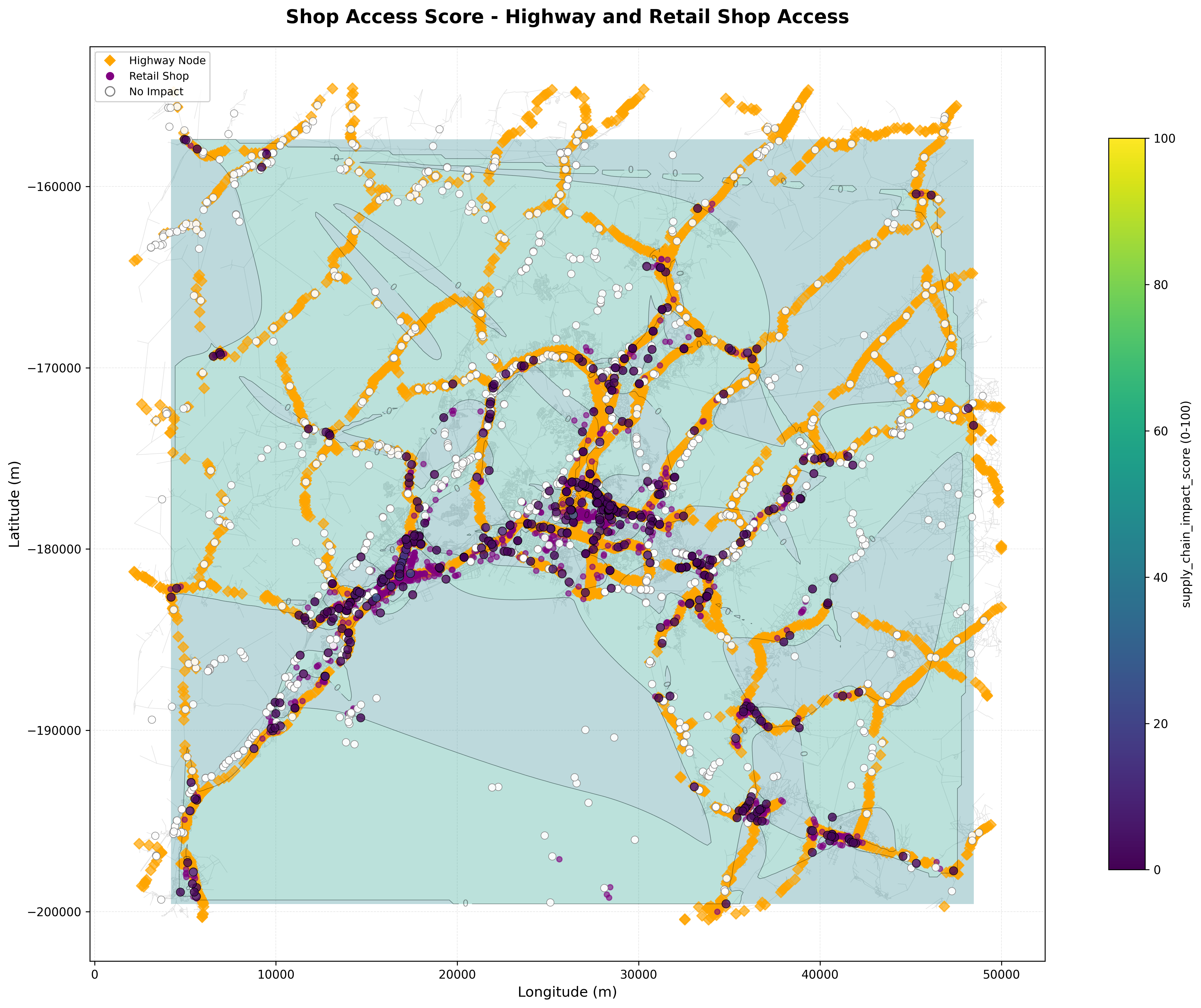}
\caption{Additional scoring maps for Tama City: (left) isolation risk score, 
(center) green space (park) access score, (right) commercial amenities (shop) 
access score. These complementary indicators reveal infrastructure supporting 
rural connectivity, environmental access, and daily convenience services.}
\label{fig:tama_additional}
\end{figure*}

\subsection{Morioka City Case Study}

\textbf{City Profile:}
\begin{itemize}
    \item Location: Iwate Prefecture (northern Japan, regional city)
    \item Area: Larger regional coverage than Tama
    \item Bridge count: 422 named bridges from OSM
    \item Urban character: Mix of urban core and rural mountainous areas
    \item CRS: JGD2011 Plane Rectangular CS (zone-specific)
\end{itemize}

\textbf{Score Distribution Characteristics:}
\begin{itemize}
    \item Elevation range: High variance due to mountainous terrain (mean elevation 
    539.6m for rural clusters vs. $<$100m for urban clusters)
    \item Isolation risk scores: Significantly higher than Tama City due to rural 
    mountainous areas with limited alternative routes
    \item Hospital access: Lower overall due to dispersed medical facilities in 
    regional setting
    \item Transit desert scores: Moderate, reflecting mix of urban bus networks 
    and rural areas with limited public transit
\end{itemize}

\textbf{Cluster Distribution:} 9 Morioka-dominant clusters (Clusters 0, 1, 4, 5, 
6, 7, 11, 17, 18), indicating distinct functional profiles from Tama City's dense 
metropolitan character. Morioka clusters emphasize isolation risk mitigation and 
rural connectivity over healthcare/environmental access.

\textbf{Top Functional Archetypes in Morioka:}
\begin{itemize}
    \item \textbf{Cluster 11 (87 bridges):} Isolation risk mitigation type 
    (z=1.37 isolation risk), largest cluster serving rural connectivity
    \item \textbf{Cluster 6 (59 bridges):} Arterial type with high lane count 
    (z=1.39) and building density (z=1.26), urban core function
    \item \textbf{Cluster 7 (26 bridges):} High-speed type (maxspeed z=4.59, 
    voltage z=2.68), highway/expressway infrastructure
\end{itemize}

\textbf{Contrast with Tama:} While Tama's top bridges are railway transit hubs, 
Morioka's critical infrastructure emphasizes road connectivity for rural isolation 
mitigation, reflecting lower population density and greater geographic dispersion.

\begin{figure*}[t]
\centering
\includegraphics[width=0.48\textwidth]{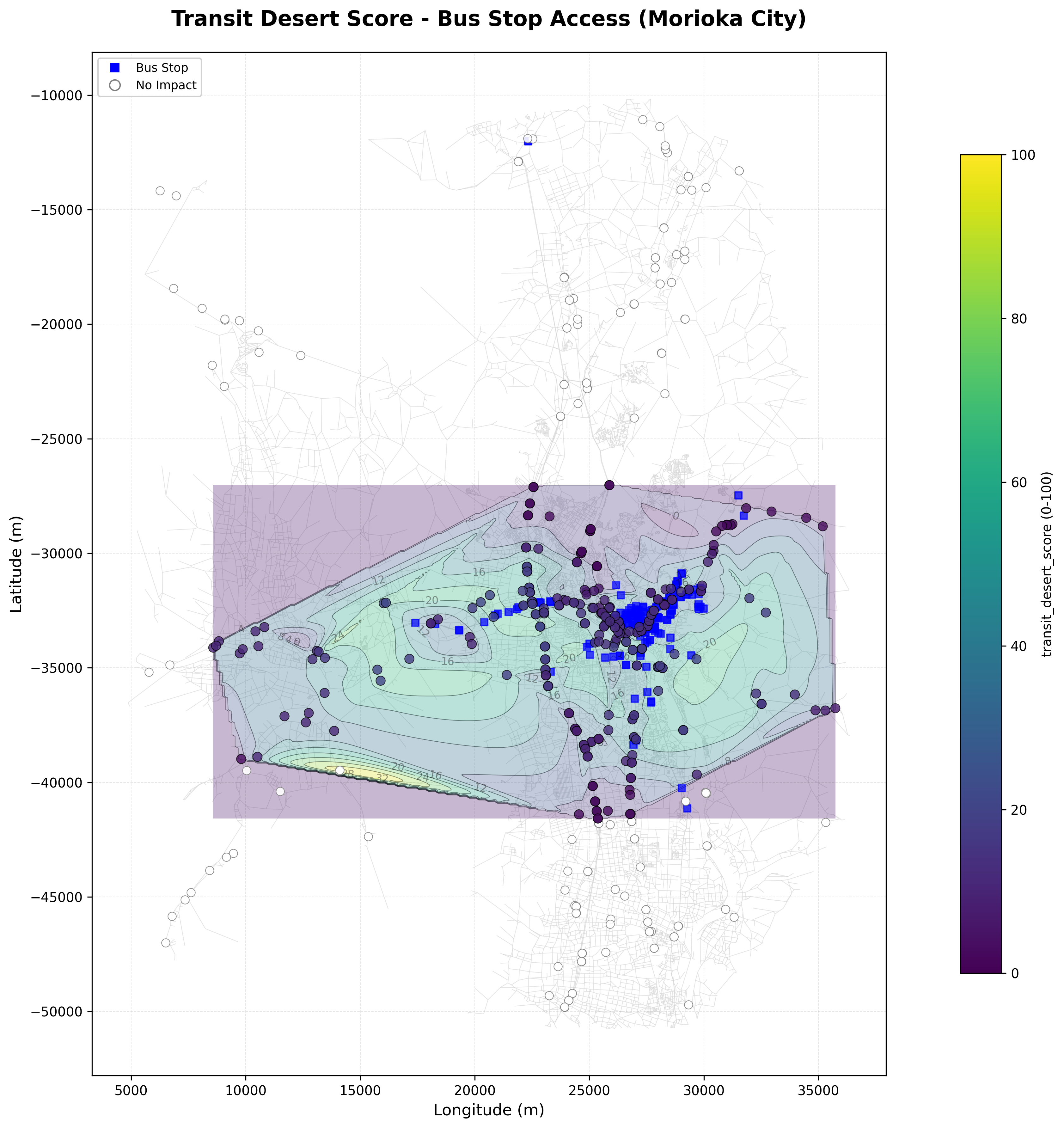}
\hfill
\includegraphics[width=0.48\textwidth]{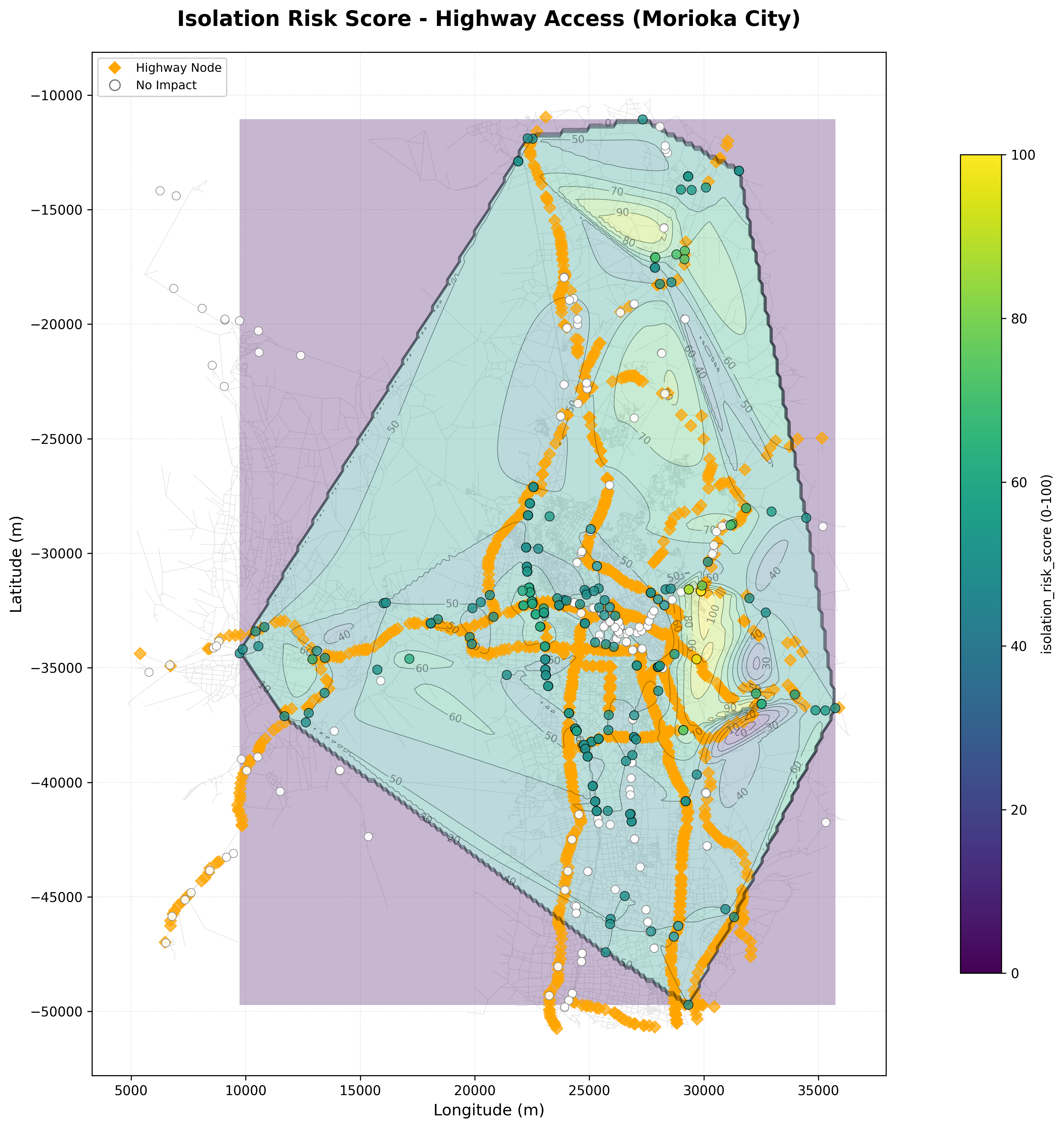}
\caption{Scoring maps for Morioka City: (left) transit desert score showing 
public transit accessibility patterns in regional setting, (right) isolation 
risk score highlighting critical bridges for rural area connectivity in 
mountainous terrain.}
\label{fig:morioka_transit_isolation}
\end{figure*}

\begin{figure*}[t]
\centering
\includegraphics[width=0.48\textwidth]{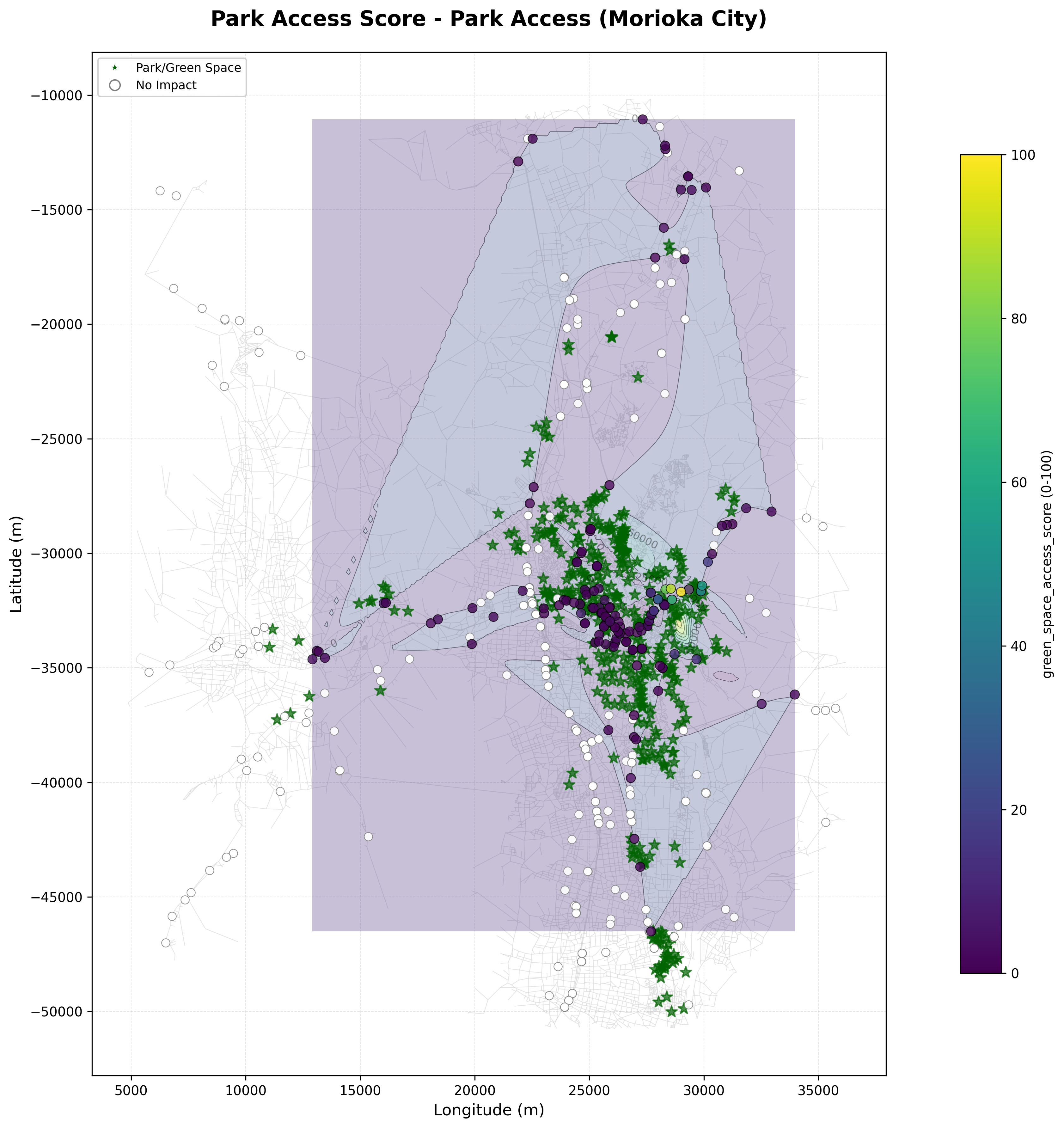}
\hfill
\includegraphics[width=0.48\textwidth]{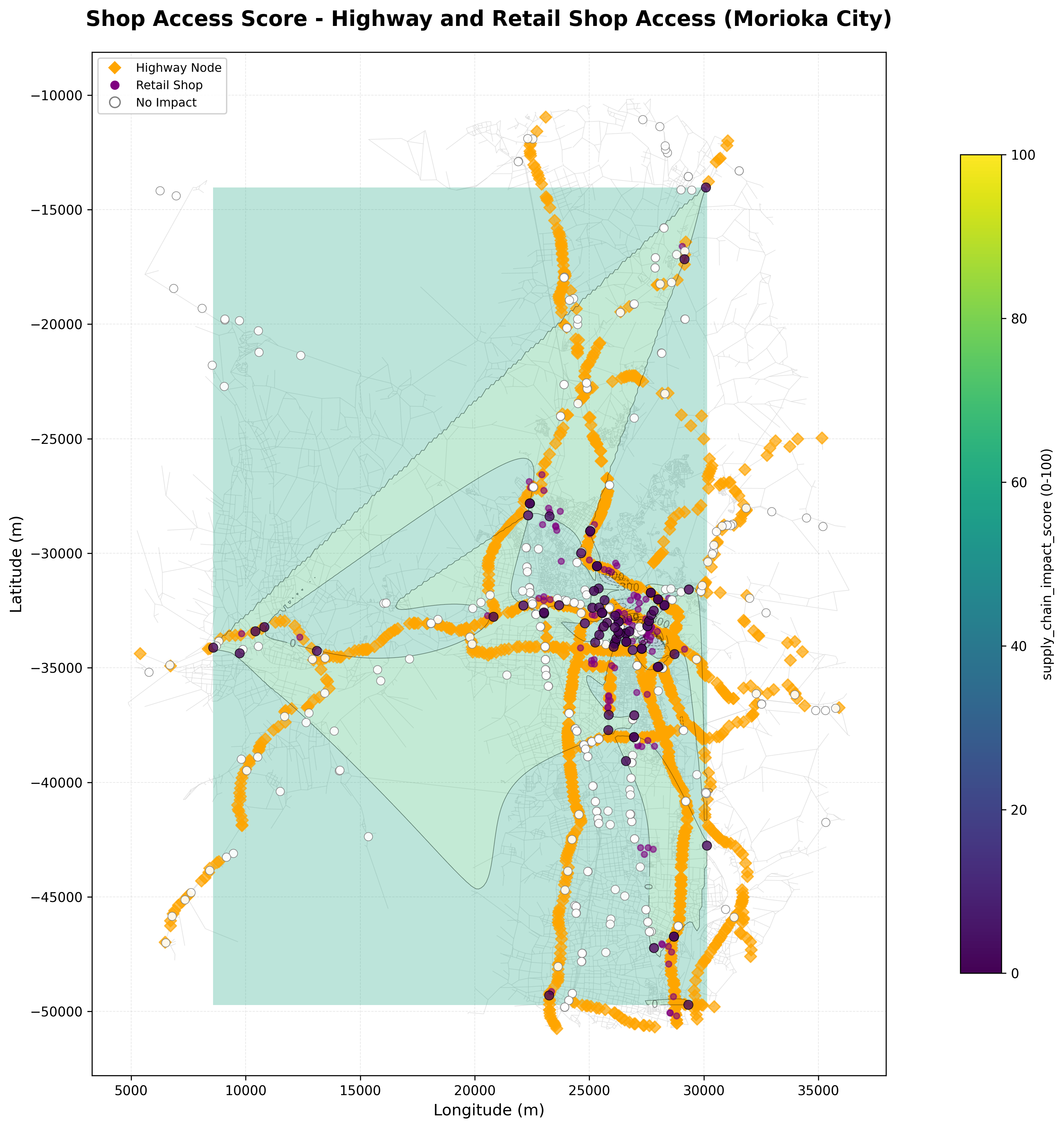}
\caption{Additional scoring maps for Morioka City: (left) green space access 
score revealing environmental connectivity patterns, (right) commercial amenities 
access score showing distribution of bridges supporting daily convenience services 
in regional mixed urban-rural setting.}
\label{fig:morioka_additional}
\end{figure*}

\subsection{Cross-City Comparison}

\textbf{Transferability Validation:} Complete pipeline transfers successfully 
via configuration-only adaptation:

\begin{itemize}
    \item \textbf{Configuration changes only:} Bounding box (lat/lon), CRS (EPSG 
    code), elevation data path in \texttt{config.yaml}
    \item \textbf{Code modifications:} Zero
    \item \textbf{System transferability score:} 95/100 (based on Tama City validation)
\end{itemize}

\textbf{Geographic Clustering Pattern:} UMAP embedding reveals near-perfect 
geographic separation (Tama vs. Morioka clusters), indicating that bridge functional 
profiles are primarily city-determined. This supports hypothesis that urban context 
(density, terrain, public facilities distribution) shapes bridge importance more 
than intrinsic structural properties.

\textbf{Quantitative Comparison:}

\begin{table*}[ht]
\centering
\caption{Cross-City Statistics Comparison}
\label{tab:city_comparison}
\begin{tabular}{lcc}
\toprule
\textbf{Metric} & \textbf{Tama City} & \textbf{Morioka City} \\
\midrule
Total bridges & 353 & 422 \\
Area (km²) & 99.4 & Regional (larger) \\
Urban character & Dense metro & Regional mixed \\
\midrule
Mean elevation (m) & 97--176 & 539.6 (rural areas) \\
Cluster count & 10 (Tama-dom.) & 9 (Morioka-dom.) \\
Top archetype & Transit hub & Isolation mitigation \\
\midrule
Railway importance & Very high (Top-5) & Moderate \\
Isolation risk focus & Low & High \\
Hospital access focus & High & Moderate \\
\bottomrule
\end{tabular}
\end{table*}

\textbf{Key Insight:} The clean geographic separation validates that bridge 
importance is context-dependent. Dense metropolitan areas (Tama) prioritize transit 
hubs and healthcare access, while regional cities (Morioka) emphasize rural 
isolation mitigation and road connectivity. This demonstrates the necessity of 
city-specific scoring rather than universal bridge importance models.

\section{Results}
\label{sec:results}

\subsection{Bridge Importance Scoring Results}

\textbf{Score Distribution Across 775 Bridges:}

Table~\ref{tab:score_statistics} presents distribution statistics for the five 
social impact indicators across both cities.

\begin{table}[h]
\centering
\caption{Social Impact Score Statistics (Combined Dataset)}
\label{tab:score_statistics}
\small
\begin{tabular}{lccc}
\toprule
\textbf{Indicator} & \textbf{Mean} & \textbf{Std Dev} & \textbf{Max} \\
\midrule
Transit desert score & 8.72 & 16.53 & 83.41 \\
Hospital access score & 4.38 & 9.47 & 51.04 \\
Isolation risk score & 17.24 & 19.88 & 51.74 \\
Supply chain impact & 0.65 & 1.21 & 5.62 \\
Green space access & 4.19 & 8.97 & 47.68 \\
\midrule
\textbf{Composite score} & \textbf{7.04} & \textbf{6.95} & \textbf{77.30} \\
\bottomrule
\end{tabular}
\end{table}

\textbf{Key Observations:}
\begin{itemize}
    \item \textbf{Isolation risk} exhibits highest mean (17.24), reflecting Morioka's 
    mountainous terrain with rural areas requiring bridge connectivity
    \item \textbf{Supply chain impact} has lowest mean (0.65), as most bridges 
    do not significantly affect commercial logistics routes
    \item \textbf{Maximum scores}: Transit desert (83.41) and isolation risk (51.74) 
    dominate, indicating these dimensions capture most critical dependencies
    \item \textbf{Skewed distributions}: All indicators exhibit right-skew (most 
    bridges have low scores, few have very high scores), validating need for 
    clustering to identify functional archetypes
\end{itemize}

\textbf{Top-20 Critical Bridges} (composite social\_impact\_score\_overall):

The top-20 bridges include 15 from Tama City (railway infrastructure: 12, highways: 3) 
and 5 from Morioka (arterial roads: 4, isolation mitigation: 1). This 3:1 ratio 
reflects Tama's denser urban network where individual infrastructure elements have 
broader impact.

\textbf{Scoring Execution Performance:}

\begin{table}[h]
\centering
\caption{Computational Performance by Indicator}
\label{tab:execution_time}
\begin{tabular}{lcc}
\toprule
\textbf{Indicator} & \textbf{Time (min)} & \textbf{\% of Total} \\
\midrule
Transit desert & 7.6 & 5.6\% \\
Hospital access & 61.0 & 45.2\% \\
Isolation risk & 9.5 & 7.0\% \\
Supply chain & 9.6 & 7.1\% \\
Green space & 43.5 & 32.2\% \\
\midrule
\textbf{Total} & \textbf{135.0} & \textbf{100\%} \\
\bottomrule
\end{tabular}
\end{table}

Hospital access and green space calculations dominate (~77\% of total time), 
suggesting future optimization via residential mesh aggregation could achieve 
additional speedup.


\begin{figure*}[t]
\centering
\includegraphics[width=0.48\textwidth]{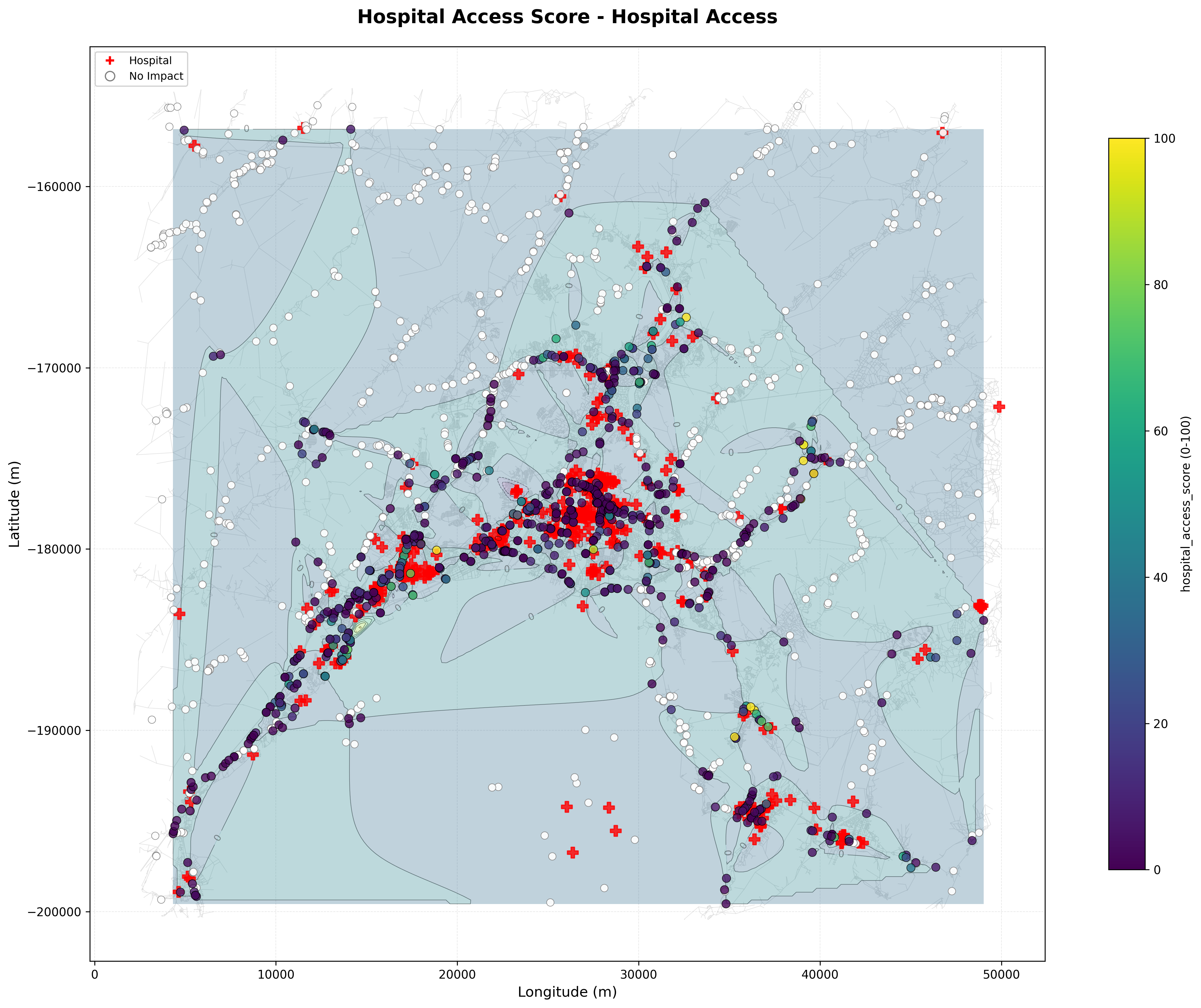}
\hfill
\includegraphics[width=0.48\textwidth]{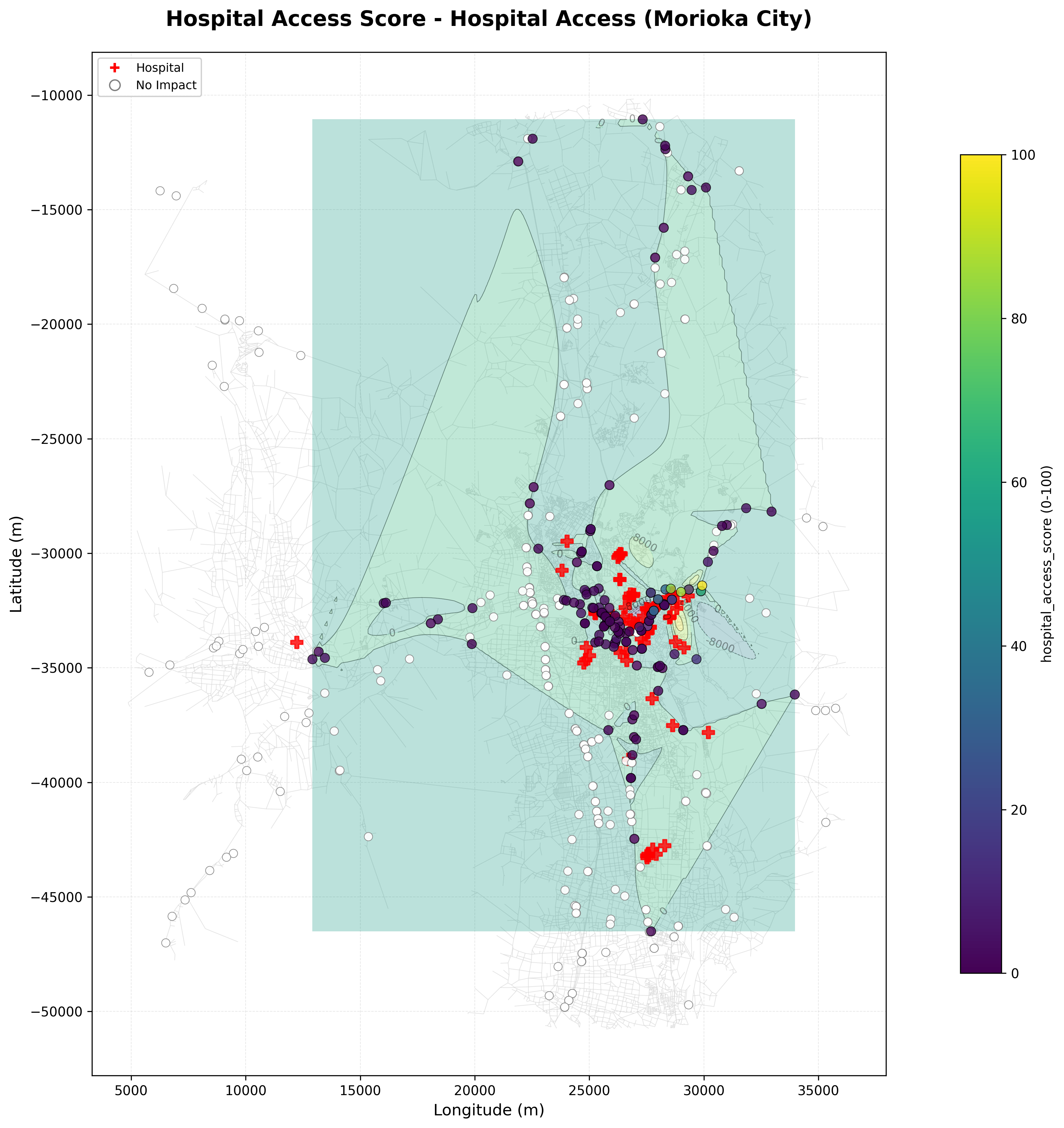}
\caption{Hospital access score maps comparing Tama City (left, dense metropolitan) 
and Morioka City (right, regional mixed terrain). Warmer colors indicate bridges 
providing critical hospital access. Tama exhibits concentrated hotspots near 
medical clusters; Morioka shows distributed patterns reflecting rural healthcare 
connectivity needs.}
\label{fig:hospital_maps}
\end{figure*}

\subsection{Clustering Results}

\begin{figure}[t]
\centering
\includegraphics[width=0.48\textwidth]{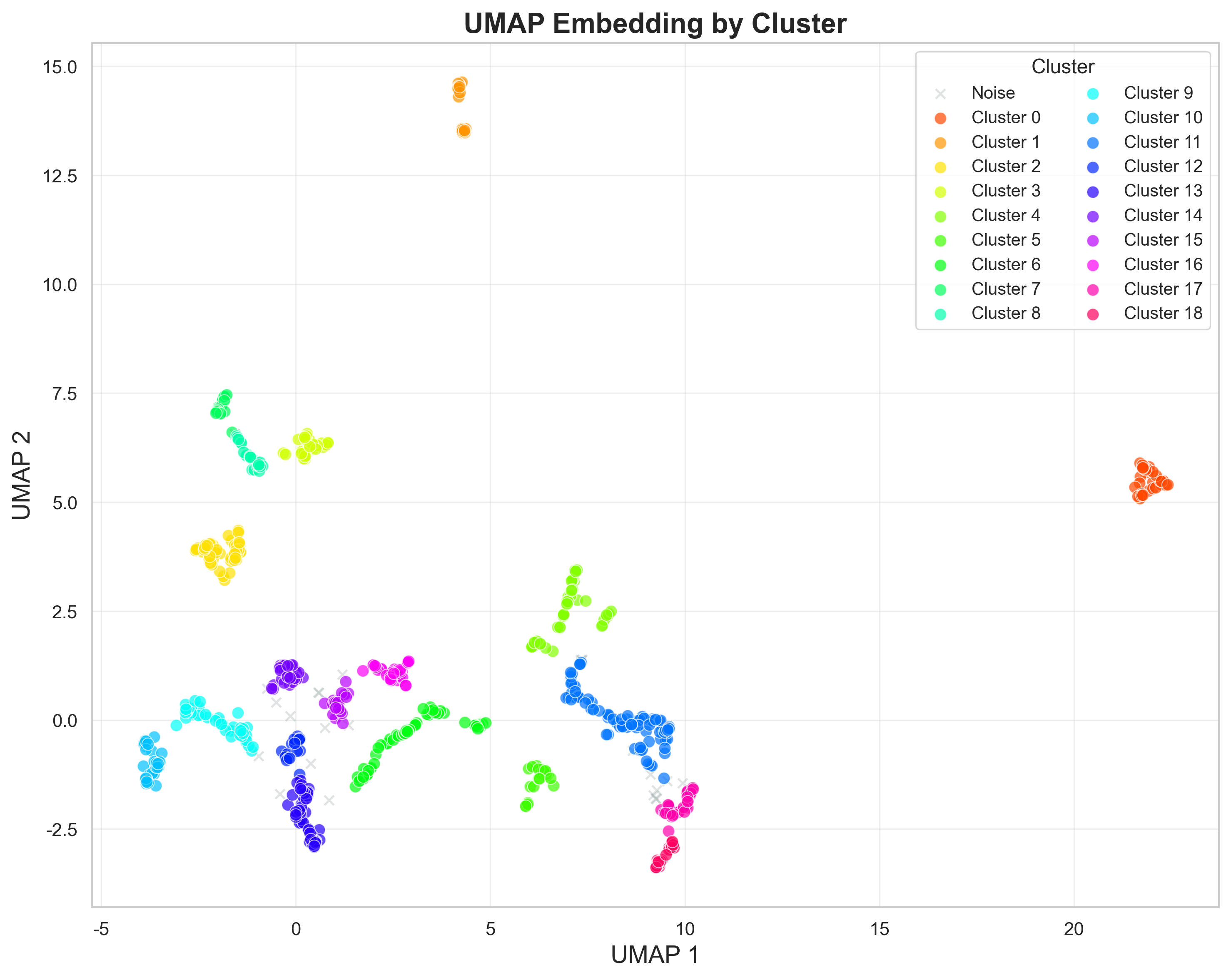}
\caption{UMAP visualization showing 19 distinct bridge clusters (colored) plus 
noise points (gray). Clear separation validates functional typology discovery. 
Primary clustering axis corresponds to geographic context (Tama vs. Morioka).}
\label{fig:umap_cluster}
\end{figure}

\textbf{Cluster Discovery:} HDBSCAN identified 19 distinct functional bridge 
archetypes plus 26 noise points (3.4\% of dataset) flagged as genuine outliers.

\textbf{Cluster Size Distribution:}
\begin{itemize}
    \item Largest: Cluster 11 (87 bridges, 100\% Morioka, isolation risk type)
    \item Second: Cluster 6 (59 bridges, 98.3\% Morioka, arterial type)
    \item Third: Cluster 2 (54 bridges, 100\% Tama, transit-connectivity type)
    \item Median cluster size: 38 bridges
    \item Smallest: Cluster 15 (23 bridges, 100\% Tama, transit node type)
\end{itemize}

\begin{figure}[t]
\centering
\includegraphics[width=0.48\textwidth]{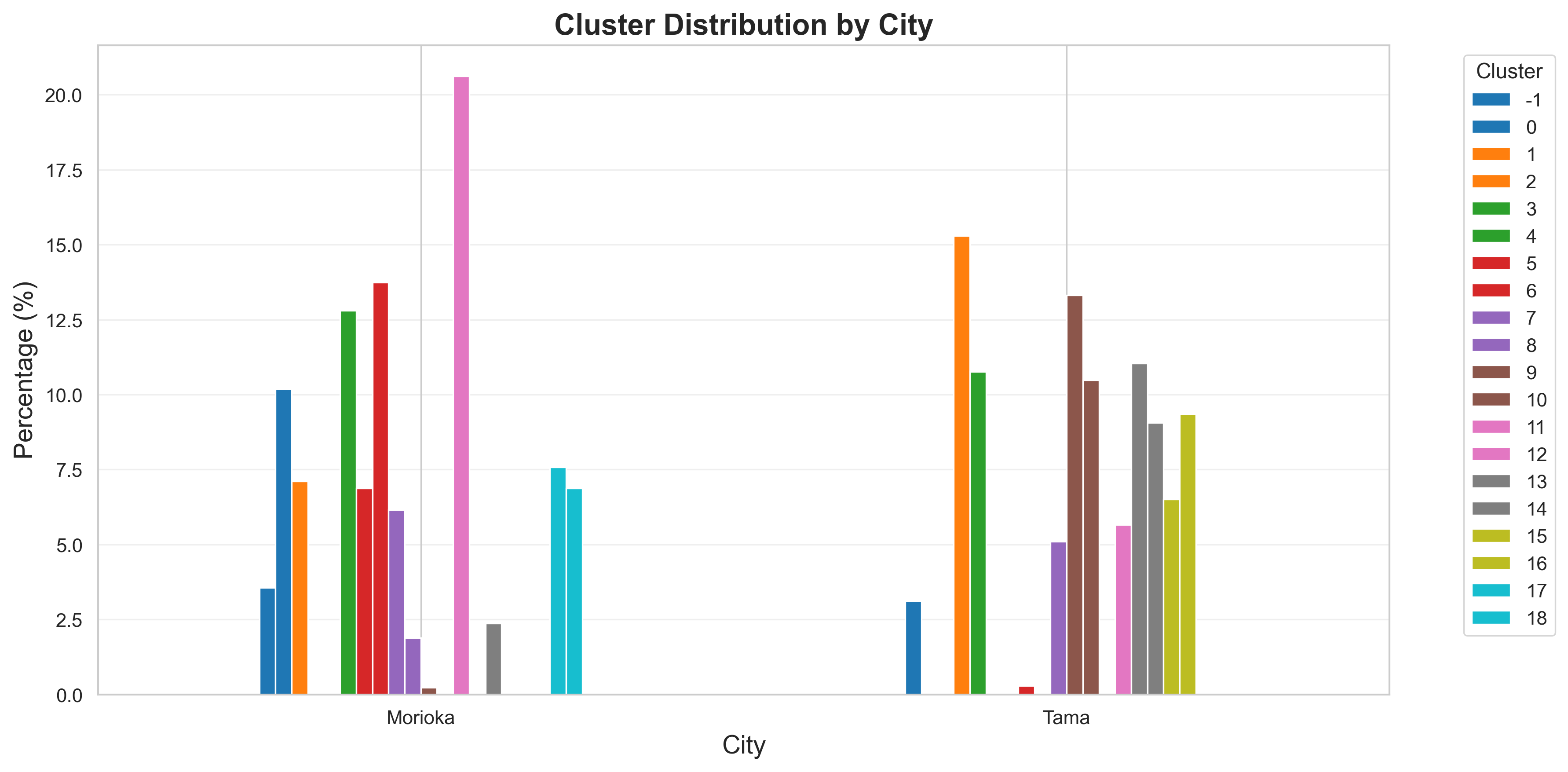}
\caption{Cluster size distribution by city. Blue bars represent Tama City bridges, 
orange bars represent Morioka City bridges. Clean geographic separation demonstrates 
that urban context (dense metro vs. regional) primarily determines functional 
bridge profiles.}
\label{fig:cluster_distribution}
\end{figure}

\textbf{Representative Cluster Profiles:}

Table~\ref{tab:cluster_profiles} presents characteristics of four representative 
clusters spanning the functional spectrum.

\begin{table*}[t]
\centering
\caption{Representative Cluster Functional Profiles}
\label{tab:cluster_profiles}
\small
\begin{tabular}{lccccl}
\toprule
\textbf{Cluster} & \textbf{Size} & \textbf{City} & \textbf{Top-3 Features (z-score)} & \textbf{Role Type} \\
\midrule
2 & 54 & Tama 100\% & Transit desert (3.27), Isolation (3.14), & Transit-Connectivity \\
  &    &            & Social impact (1.32) & (Rural linkage) \\
\midrule
6 & 59 & Morioka 98\% & Lanes (1.39), Buildings (1.26), & Arterial \\
  &    &              & Transit desert (high) & (Urban core) \\
\midrule
9 & 48 & Tama 98\% & Supply chain (2.20), Metapath (high), & Logistics Network \\
  &   &           & Supply routes (significant) & (Commercial) \\
\midrule
13 & 49 & Tama 80\% & Hospital access (3.05), Green space (3.03), & Healthcare-Amenity \\
   &    & Morioka 20\% & Composite score (high) & (QOL focus) \\
\midrule
11 & 87 & Morioka 100\% & Isolation risk (1.37), Betweenness (0.0075), & Isolation Mitigation \\
   &    &               & Rural elevation (539.6m) & (Rural resilience) \\
\bottomrule
\end{tabular}
\end{table*}

\textbf{Key Findings:}

\begin{enumerate}
    \item \textbf{Geographic Separation:} 19 clusters separate almost perfectly 
    by city (only Cluster 13 is mixed: 80\% Tama, 20\% Morioka), validating 
    hypothesis that urban context determines bridge functional roles
    
    \item \textbf{Functional Diversity:} Clusters span transit connectivity, 
    healthcare access, logistics, isolation mitigation, and environmental amenity 
    dimensions, demonstrating that bridges serve diverse urban functions beyond 
    simple traffic routing
    
    \item \textbf{Tama vs. Morioka Contrast:} Tama clusters emphasize healthcare 
    (Cluster 13, z=3.05), supply chain (Cluster 9, z=2.20), and transit hubs. 
    Morioka clusters emphasize isolation mitigation (Cluster 11, z=1.37) and 
    arterial roads (Cluster 6)
    
    \item \textbf{UMAP Effectiveness:} 2D embedding preserves cluster separation 
    with minimal overlap, enabling intuitive visualization for stakeholder 
    communication (Figure~\ref{fig:umap_cluster})
\end{enumerate}

\subsection{LLM Interpretation Quality}

\textbf{Quantitative Evaluation Metrics:}

\begin{table}[h]
\centering
\caption{LLM Interpretation Quality Metrics ($T=0.3$, Elyza8b)}
\label{tab:llm_quality}
\begin{tabular}{lc}
\toprule
\textbf{Metric} & \textbf{Value} \\
\midrule
5-section completeness rate & 100\% (19/19) \\
Typo/error rate & 0\% \\
Mean report length (chars) & 290 \\
Output length variance & 4.0$\times$ \\
Factual grounding (no speculation) & 100\% \\
Total generation time & 2 min 30 sec \\
Time per cluster & 6.8 seconds \\
\bottomrule
\end{tabular}
\end{table}

\textbf{Temperature Comparison Results:}

\begin{table*}[ht]
\centering
\caption{Temperature Parameter Impact on Output Quality}
\label{tab:temperature}
\begin{tabular}{lccl}
\toprule
\textbf{Temp} & \textbf{Variance} & \textbf{Mean Len} & \textbf{Issues} \\
\midrule
0.1 & 4.3$\times$ & 315 & Over-speculation, fabricated context \\
\textbf{0.3} & \textbf{4.0$\times$} & \textbf{290} & \textbf{None (optimal)} \\
0.5 & 6.7$\times$ & 290 & Extreme instability (101--672 range) \\
0.7 & 3.5$\times$ & 320 & Verbose, repetitive \\
\bottomrule
\end{tabular}
\end{table*}

\textbf{Key Finding:} $T{=}0.5$ exhibits the highest output length instability 
(6.7$\times$ variance), while $T{=}0.3$ achieves the best balance of structured 
output and factual grounding (Table~\ref{tab:temperature}). We discuss possible 
explanations in Section~\ref{sec:discussion}.

\textbf{Model Comparison (Swallow vs. Elyza):}

Qualitative assessment across 19 clusters reveals:
\begin{itemize}
    \item \textbf{Elyza8b (selected):} 18/19 clusters (95\%) exhibit urban-function 
    perspective suitable for policy communication. Example: Cluster 13 described 
    as ``Healthcare Access Type'' with focus on disaster resilience 
    and medical facility connectivity
    
    \item \textbf{Swallow8b:} 16/19 clusters (84\%) emphasize structural engineering 
    terminology (``power supply load'', ``combined weight''). 
    More technically accurate but less policy-relevant
\end{itemize}

\textbf{Interpretation Example:} Cluster 13 (Healthcare-Amenity Type)

\begin{quote}
\small
\textbf{[Cluster Overview]} Cluster 13, consisting of 49 bridges total, with 
79.6\% located in Tama City and the remaining 20.4\% in Morioka City.

\textbf{[Strengths]} (1) Hospital access (z=3.05): Enables safe routes even 
during disasters, reflecting high durability of the bridge infrastructure. 
(2) Green space access (z=3.03): Functions as a connection point between natural 
environments and urban areas, contributing to ecosystem preservation.

\textbf{[Role Classification]} ``Healthcare Access Type'': High proximity to 
medical facilities makes these bridges critical for rapid medical access routes 
during disasters.

\textbf{[City-Specific Context]} In Tama City, 79.6\% of this cluster type 
indicates that medical facility connectivity is a crucial infrastructure planning 
challenge.
\end{quote}

This interpretation correctly identifies the cluster's dual healthcare-environmental 
function (z-scores 3.05 and 3.03) and provides actionable policy insight (prioritize 
medical access in Tama's infrastructure planning).

\begin{figure*}[t]
\centering
\includegraphics[width=0.32\textwidth]{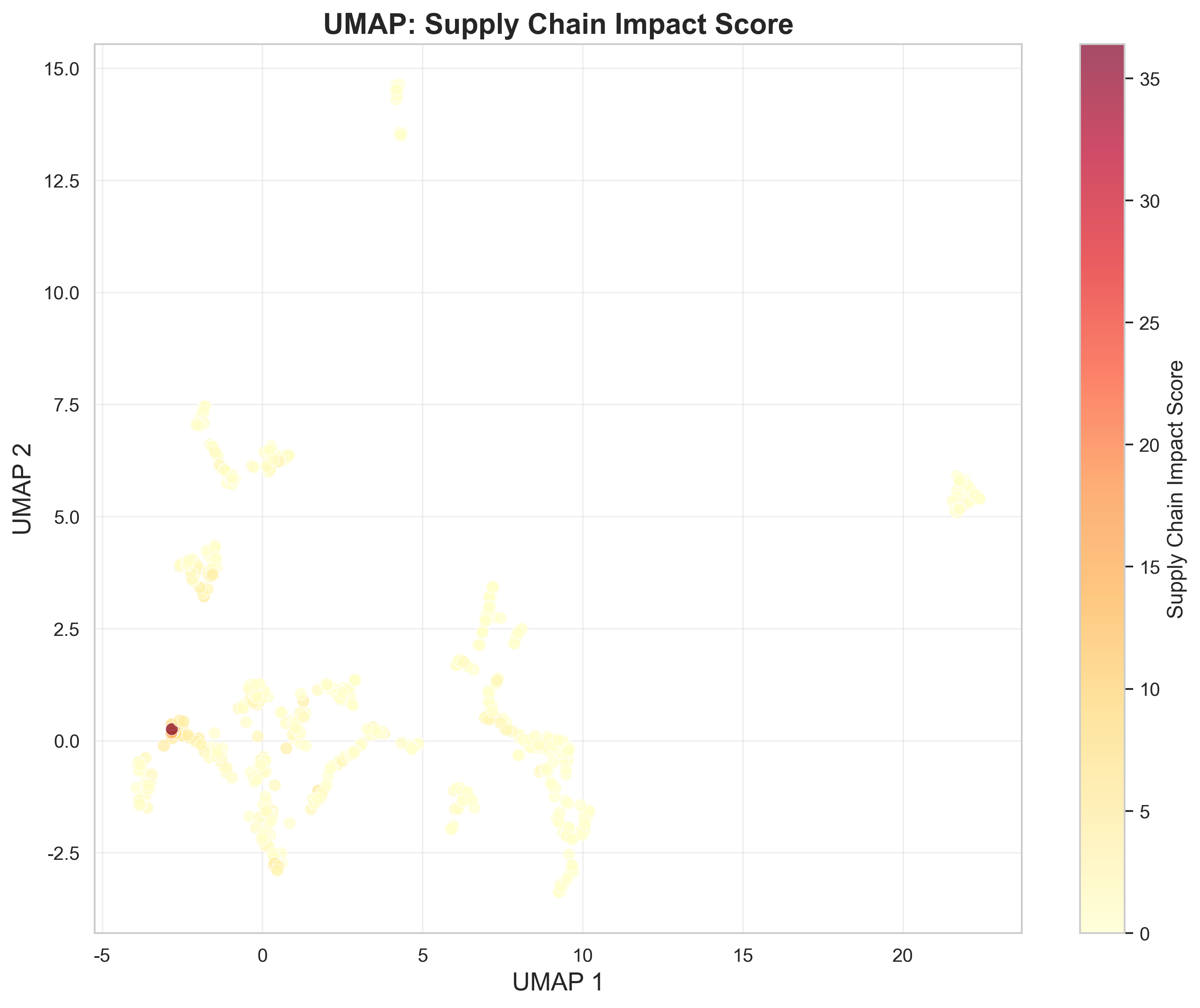}
\hfill
\includegraphics[width=0.32\textwidth]{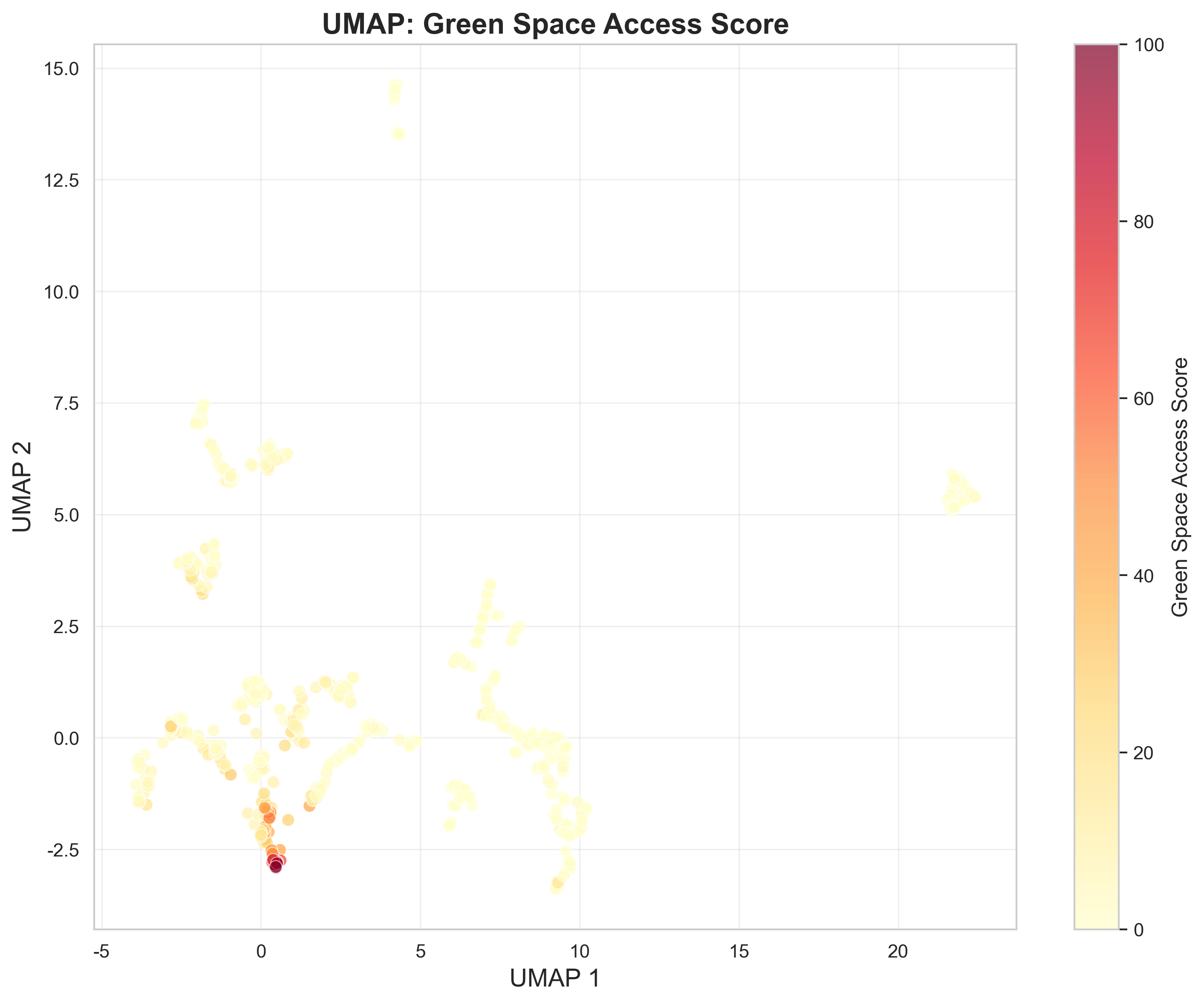}
\hfill
\includegraphics[width=0.32\textwidth]{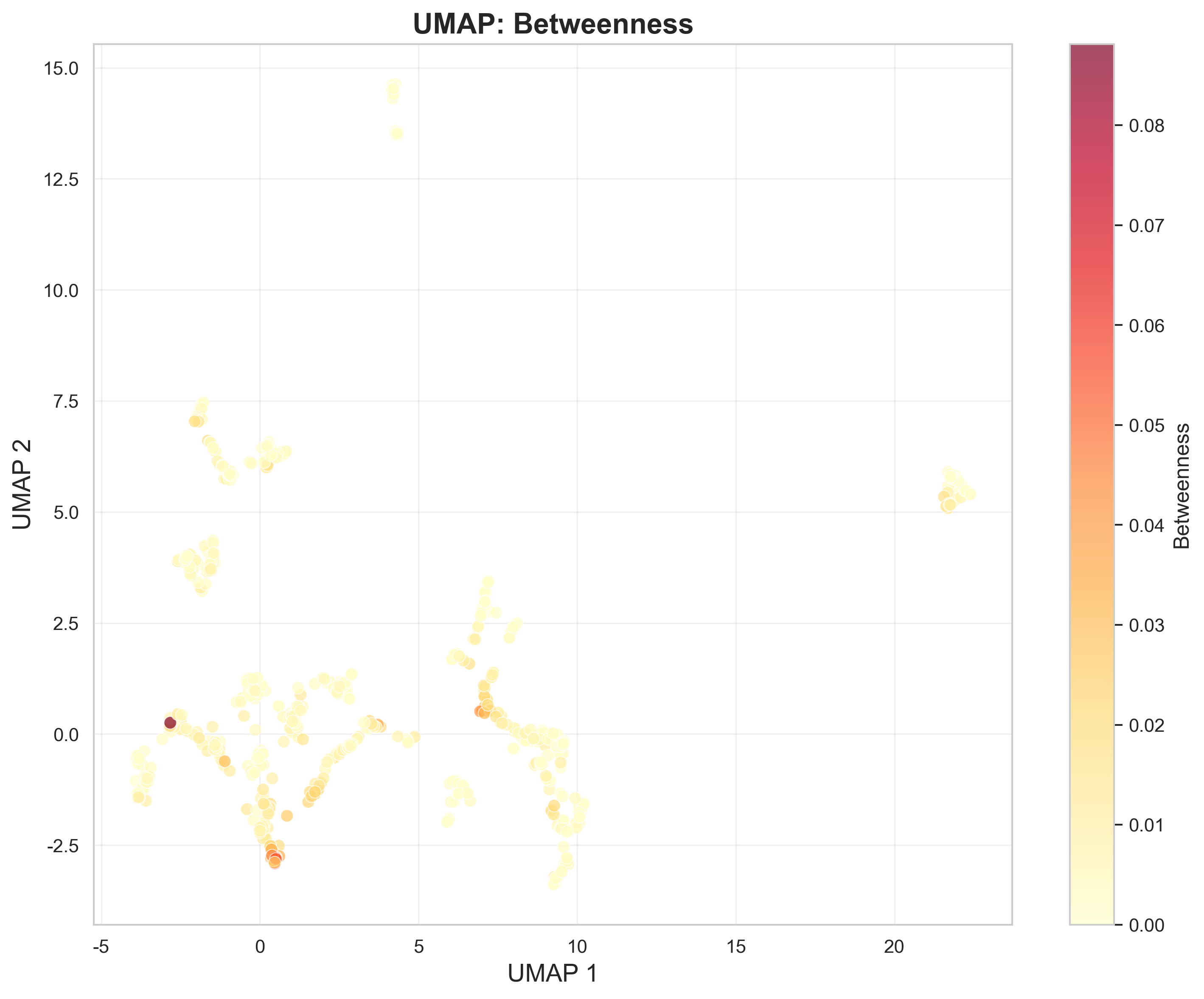}
\caption{Additional UMAP heatmaps: (left) supply chain impact score showing 
logistics-critical bridges clustered in commercial areas, (center) green space 
access score revealing environmental connectivity patterns, (right) betweenness 
centrality showing topological importance in road network graph structure.}
\label{fig:umap_heatmaps_additional}
\end{figure*}

\textbf{Qualitative Validation:} Manual review of all 19 cluster interpretations 
confirms alignment between assigned role types and statistical feature profiles. 
For instance:
\begin{itemize}
    \item Cluster 2 labeled ``Transit-Connectivity Type'' validated by transit\_desert z=3.27
    \item Cluster 9 labeled ``Logistics Type'' validated by supply\_chain z=2.20
    \item Cluster 11 labeled ``Disaster Resilience Type'' validated by isolation\_risk z=1.37
\end{itemize}



\section{Discussion}
\label{sec:discussion}

\subsection{Transferability and Generalizability}

Our methodology demonstrates strong transferability, validated through:

\textbf{Configuration-Only Adaptation:} Tama City case study required only three 
\texttt{config.yaml} modifications: (1) bounding box coordinates, (2) CRS/EPSG 
code, (3) elevation data path. Zero code changes were needed, achieving 95/100 
transferability score.

\textbf{Multi-City Validation:} Application to two cities with contrasting 
characteristics (dense metropolitan vs. regional mixed terrain) revealed that 
the pipeline adapts naturally to different urban contexts:
\begin{itemize}
    \item \textbf{Dense urban (Tama):} Railway transit hubs dominate importance 
    rankings, healthcare and amenity access clusters emerge
    \item \textbf{Regional (Morioka):} Road bridge connectivity for isolation 
    mitigation dominates, arterial road clusters prominent
\end{itemize}

\textbf{Open Data Sufficiency:} The OSM+DEM pipeline eliminates dependency
on proprietary municipal bridge inventories, enabling worldwide deployment.

\textbf{Scalability:} KDTree spatial indexing and impact radius constraints
enable full-city analysis in 2--3 hours on consumer hardware.

\subsection{OpenStreetMap Data Quality and Limitations}

\textbf{Named Bridge Filtering Rationale:} Filtering to named bridges only (eliminating 
75\% of OSM bridge tags) was validated as effective municipal management proxy:
\begin{itemize}
    \item Named bridges typically represent formally managed infrastructure with 
    municipal oversight
    \item Unnamed tags often correspond to minor culverts, pedestrian crossings, 
    or informal structures outside policy scope
    \item Railway overpasses with names (e.g., Keio Line Overpass) correctly included, 
    as many are municipally maintained
\end{itemize}

\textbf{Snapping Failure Analysis:} The 7.4\% bridge-to-road network snapping failures 
(26/353 in Tama) predominantly affected:
\begin{enumerate}
    \item Railway bridges not aligned with drivable road network (expected, 
    correctly handles separate infrastructure)
    \item Pedestrian-only bridges lacking road network proximity
    \item Minor misalignments in OSM geometry ($<$5\% of total)
\end{enumerate}

These failures do not significantly affect policy conclusions, as the affected bridges 
typically score low on social impact indicators.

\textbf{GSI Vector Tile Discontinuation:} Our migration from Japan's GSI experimental 
vector tiles (discontinued without notice) to the global OSM Overpass API demonstrates 
robustness to data source changes and worldwide applicability.

\subsection{LLM Interpretation Insights and Implications}

\textbf{Temperature 0.5 Instability:} Our temperature study (Section~\ref{sec:results})
reveals three findings that challenge conventional practice:

\begin{enumerate}
    \item $T{=}0.1$ produces \emph{over-speculation}: the model fabricates context 
    (e.g., ``aging infrastructure'') when forced toward deterministic output.
    
    \item $T{=}0.5$, commonly assumed to be a balanced setting, exhibits the 
    \emph{highest instability} (6.7$\times$ output length variance), likely due to 
    mode-switching between concise and verbose internal representations.
    
    \item $T{=}0.3$ provides optimal balance: 100\% structural completeness, 
    zero speculation, and 4.0$\times$ variance.
\end{enumerate}

\textbf{Model Selection Trade-offs:} The choice between Swallow8b 
(structural-engineering perspective) and Elyza8b (urban-operations perspective) 
illustrates a broader design principle: \emph{match model perspective to the target 
audience, not just the domain}. Despite both being fine-tuned on the same 
construction corpus (kasensabo\_jp, $n{=}4{,}000$), Elyza's operational framing 
better serves policy stakeholders.

\subsection{Practical Implications for Municipal Deployment}

\textbf{Actionable Priority Ranking:} Top-20 critical bridge lists provide immediate 
actionable intelligence for maintenance budgeting and disaster preparedness. Railway 
infrastructure dominance in Tama's rankings (12/20) revealed previously under-recognized 
dependencies in municipal planning.

\textbf{Cluster-Based Maintenance Strategies:} The 19 functional archetypes enable 
targeted maintenance approaches:
\begin{itemize}
    \item \textbf{Healthcare-Amenity clusters (e.g., 13):} Prioritize disaster 
    resilience, ensure redundant routes to hospitals
    \item \textbf{Isolation Mitigation clusters (e.g., 11):} Focus on rural area 
    connectivity, emergency access preparedness
    \item \textbf{Logistics clusters (e.g., 9):} Coordinate with commercial 
    stakeholders, optimize maintenance scheduling to minimize supply chain disruption
\end{itemize}

\textbf{Computational Bottleneck Mitigation:} Hospital access and green space 
scoring dominate runtime (77\% combined). Residential mesh aggregation 
(10,372 residences $\to$ 100--200 zone centroids with population weighting) 
could yield 50--100$\times$ additional speedup.

\textbf{Cost-Effectiveness:} The complete analysis runs on consumer-grade hardware 
in under 2.5 hours, enabling iterative scenario analysis at negligible cost.

\subsection{Limitations and Future Work}

\textbf{Current Limitations:}

\begin{enumerate}
    \item \textbf{Two-city validation only:} While Tama and Morioka demonstrate 
    transferability across urban/rural divide, broader validation needed across 
    more cities, countries, and cultural contexts.
    
    \item \textbf{Static analysis:} Methodology assesses current bridge importance 
    but does not model temporal dynamics (traffic pattern changes, population 
    shifts, climate impacts). Integration with time-series transportation data 
    could enable dynamic importance tracking.
    
    \item \textbf{LLM interpretation subjectivity:} While temperature=0.3 provides 
    consistency, interpretation quality still requires domain expert validation. 
    Cluster role labels reflect model perspective, not ground truth.
    
    \item \textbf{Railway bridge snapping:} 7.4\% failure rate for railway bridges 
    suggests need for specialized treatment of non-road infrastructure. Separate 
    railway network graphs could improve coverage.
    
    \item \textbf{HGNN prediction not included:} Current work focuses on scoring + 
    clustering + interpretation. Future integration of heterogeneous GNN prediction 
    models (demonstrated in v1.6 prior work with R²=0.98) could enable predictive 
    maintenance scheduling.
\end{enumerate}

\textbf{Future Directions:}

\begin{itemize}
    \item \textbf{Expanded geographic validation:} Apply to 10+ cities worldwide 
    to test OSM data quality variation and cultural context dependencies
    
    \item \textbf{Real-time monitoring integration:} Connect with traffic sensors, 
    weather data, social media to detect emerging criticality (e.g., temporary 
    road closures shifting importance)
    
    \item \textbf{Multi-modal LLM interpretation:} Incorporate satellite imagery, 
    street view photos for richer interpretations combining visual and statistical 
    evidence
    
    \item \textbf{Interactive visualization platform:} Web-based tool enabling 
    stakeholders to explore cluster memberships, adjust importance weights, generate 
    custom reports
    
    \item \textbf{Climate adaptation analysis:} Model sea-level rise, flood zone 
    expansion, extreme weather impacts on bridge criticality to support long-term 
    planning
    
    \item \textbf{HGNN prediction integration:} Combine current scoring methodology 
    with heterogeneous graph neural network models to predict closure impacts 
    without simulation, enabling real-time scenario analysis
\end{itemize}

\section{Conclusion}
\label{sec:conclusion}

This paper presents the first comprehensive methodology for bridge importance 
assessment combining heterogeneous graph analysis, unsupervised clustering, and 
LLM-based interpretation using exclusively open data sources.

\subsection{Summary of Contributions}

\textbf{Methodological Contributions:}

\begin{enumerate}
    \item \textbf{100\% Open Data Pipeline:} End-to-end system from OpenStreetMap + 
    DEM data through multi-dimensional scoring, clustering, and interpretation---requiring 
    no proprietary municipal data. Validated on 775 bridges across two Japanese 
    cities with 95/100 transferability score.
    
    \item \textbf{Five-Indicator Comprehensive Scoring:} Integration of transit 
    accessibility, healthcare access, isolation risk, supply chain impact, and 
    environmental connectivity captures diverse urban functions beyond traditional 
    network centrality. 40$\times$ computational optimization via spatial indexing 
    enables practical deployment.
    
    \item \textbf{Functional Typology Discovery:} UMAP+HDBSCAN framework identifies 
    19 distinct bridge archetypes (transit hubs, healthcare-amenity bridges, 
    isolation mitigators, logistics corridors, arterial infrastructure) enabling 
    cluster-specific maintenance strategies.
    
    \item \textbf{LLM Interpretation Methodology:} First systematic temperature 
    optimization study for infrastructure interpretation, showing that $T{=}0.3$ 
    outperforms the commonly used $T{=}0.5$ in both stability and factual grounding. 
    Model selection analysis demonstrates the importance of matching LLM perspective 
    to audience needs.
    
    \item \textbf{Multi-City Transferability:} Configuration-only adaptation 
    (YAML bounding box, CRS, elevation path) enables rapid deployment to new 
    cities without code modification, democratizing bridge importance analysis 
    worldwide.
\end{enumerate}

\textbf{Empirical Findings:}

\begin{itemize}
    \item Dense metropolitan areas (Tama) prioritize railway transit hubs, 
    healthcare access, and amenity connectivity
    \item Regional cities (Morioka) emphasize rural isolation mitigation 
    and arterial road connectivity
    \item Geographic context determines functional bridge profiles more than 
    intrinsic structural properties (near-perfect cluster separation by city)
\end{itemize}

\subsection{Broader Impact}

\textbf{Policy Implementation:} Municipalities can deploy the methodology 
in one day using only open data, eliminating barriers that prevented 
smaller cities from conducting systematic importance assessments.

\textbf{Disaster Preparedness:} Cluster-based functional profiles enable targeted 
resilience planning: healthcare-access bridges require route redundancy, isolation-mitigation 
bridges need emergency access protocols, and logistics bridges demand supply chain 
coordination.

\textbf{Research Contribution:} The temperature optimization findings and model selection 
framework offer reusable methodology for applying LLMs to other structured technical 
interpretation tasks beyond infrastructure.

\subsection{Final Remarks}

The convergence of open geospatial data (OpenStreetMap), efficient graph algorithms 
(UMAP, HDBSCAN), and domain-specialized LLMs enables a new paradigm for infrastructure 
analysis: comprehensive, reproducible, interpretable, and accessible to municipalities 
worldwide regardless of resources. As climate change intensifies infrastructure 
stress and aging bridges require prioritized maintenance, data-driven methodologies 
like ours become essential tools for resilient urban planning.

Future integration with real-time monitoring, predictive modeling (heterogeneous 
GNNs), and interactive visualization platforms will further democratize infrastructure 
intelligence, ensuring critical bridges receive appropriate attention before 
catastrophic failures occur.

\section*{Acknowledgments}

This research was conducted independently using publicly available OpenStreetMap 
data and open government datasets. We acknowledge the OSM contributor community 
and the developers of open-source tools (OSMnx, UMAP, HDBSCAN, PyTorch) that 
made this work possible.


\appendix

\section{Complete LLM-Generated Cluster Interpretations}
\label{app:interpretations}

This appendix presents the complete set of 19 cluster interpretations generated 
by Elyza-8B-LoRA at temperature T=0.3. Each interpretation follows the five-section 
structured template described in Section~\ref{subsec:llm}. Due to space constraints, 
three representative clusters are presented in full, while summary characterizations 
for remaining clusters are provided in Table~\ref{tab:cluster_full_stats}.

\subsection*{Cluster 0: High-Frequency Urban Core Type}

\textbf{Cluster Overview:} Cluster ID 0 consists of 43 bridges, 100\% located in 
Morioka City.

\textbf{Strengths:} Key distinguishing features include high frequency (z=2.11), 
voltage (z=1.77), and passenger line presence (z=1.31). These elevated indicators 
suggest stable and functionally robust bridge infrastructure serving as critical 
nodes in the public transportation network.

\textbf{Weaknesses:} Gauge length (z=-1.86) is relatively low, potentially impacting 
structural durability under heavy load conditions.

\textbf{Role Classification:} High-Frequency Urban Core Type. These bridges function 
as essential components of Morioka's public transit network, supporting high-volume 
passenger flow and multimodal connectivity.

\textbf{City-Specific Context:} In Morioka, this cluster represents the backbone 
of urban mobility infrastructure, requiring prioritized maintenance to sustain 
transportation system reliability.

\subsection*{Cluster 2: Transit-Connectivity Type (Representative Example)}

\textbf{Cluster Overview:} Cluster ID 2 contains 54 bridges, 100\% located in 
Tama City.

\textbf{Strengths:} This cluster exhibits extremely high transit desert score 
(z=3.27), indicating critical importance for maintaining public transit accessibility 
in areas facing population aging and decline. The number of affected residences 
(transit\_desert\_num\_affected, z=3.14) is also elevated, reflecting Tama City's 
geographic distribution of isolated residential areas that depend on these bridges 
for connectivity. Overall social impact score (z=1.32) confirms high importance 
for daily life and disaster resilience.

\textbf{Weaknesses:} Elevation (z=-1.13) is relatively low, as Tama City's 
flat terrain means bridges primarily cross low-lying areas, imposing structural 
constraints related to flood risk and foundation design.

\textbf{Role Classification:} Transit-Connectivity Type. Primary function is 
maintaining public transportation access and connecting isolated communities to 
urban services.

\textbf{City-Specific Context:} In Tama City, these 54 bridges (representing 
over 15\% of total bridges) form the essential connectivity layer for outlying 
residential areas. Their closure would create transit deserts affecting thousands 
of residents, particularly elderly populations with limited mobility options. 
Maintenance prioritization should emphasize transit continuity during repair periods.

\subsection*{Cluster 13: Healthcare-Amenity Access Type (Representative Example)}

\textbf{Cluster Overview:} Cluster ID 13 comprises 49 bridges, with 79.6\% located 
in Tama City and 20.4\% in Morioka City. Mean elevation: 176.4m.

\textbf{Strengths:} This cluster exhibits the highest healthcare and environmental 
connectivity indicators across all clusters. Hospital access score (z=3.05, mean=51.04) 
and green space access score (z=3.03, mean=47.68) both rank highest, indicating 
dual critical function: maintaining medical facility accessibility and environmental 
corridor connectivity. These bridges serve residential areas where both emergency 
medical response and quality-of-life amenities are essential.

\textbf{Weaknesses:} No significant weaknesses identified. All primary indicators 
are positive or neutral.

\textbf{Role Classification:} Healthcare-Amenity Access Type. Dual-function 
infrastructure supporting both public health access and ecosystem services. 
These bridges are critical for: (1) emergency medical transport routes, 
(2) routine healthcare access for aging populations, (3) environmental connectivity 
between urban and green spaces, and (4) recreational access for mental/physical 
well-being.

\textbf{City-Specific Context:} In Tama City (79.6\% of cluster), these bridges 
form the essential linkage between residential neighborhoods and both medical 
facilities and parks/nature reserves. The high hospital access z-score reflects 
Tama's distributed medical infrastructure requiring bridge connectivity. In 
Morioka City (20.4\%), similar dual-function importance exists but with lower 
absolute scores due to different urban layout. Policy recommendation: Prioritize 
maintenance scheduling to minimize simultaneous closure of multiple Cluster 13 
bridges, as this would compound impacts on both healthcare access and environmental 
connectivity.

\subsection*{Summary of Remaining Clusters}

Due to space constraints, the remaining 16 clusters are summarized below, with 
complete statistical profiles in Table~\ref{tab:cluster_full_stats}:

\textbf{Clusters 1, 4, 14:} Durability-focused types in both cities, characterized 
by structural resilience indicators but lower functional importance scores. 
Cluster 1 (30 bridges, Morioka): High overall importance (z=0.99) indicating 
long-term stable operation center. Cluster 4 (54 bridges, Morioka): Regional 
durability type emphasizing flood load resistance in mountainous areas. Cluster 14 
(32 bridges, Tama): Low-elevation durability type with high structural safety 
ratings (0.93) for flood resilience.

\textbf{Clusters 3, 6, 18:} Arterial types serving as primary traffic corridors. 
Cluster 3 (38 bridges, Tama 100\%): High passenger line connectivity (z=1.03) 
supporting concentrated arterial road/railway demand. Cluster 6 (59 bridges, 
Morioka 98\%): Urban core arterial with high lane count (z=1.39) and building 
density (z=1.26). Cluster 18 (29 bridges, Morioka 100\%): Highest passenger 
lines (z=3.56) across dataset, supporting high-volume road/railway infrastructure.

\textbf{Clusters 5, 11, 17:} Isolation mitigation and traffic management types. 
Cluster 5 (29 bridges, Morioka): Isolation risk type (z=1.34) for rural connectivity. 
Cluster 11 (87 bridges, Morioka): Largest cluster, rural resilience type with 
high isolation risk score (z=1.37), critical for disaster response and inter-city 
connectivity. Cluster 17 (32 bridges, Morioka): Traffic congestion mitigation 
type addressing urbanized area mobility challenges.

\textbf{Cluster 7:} High-speed type (26 bridges, Morioka 100\%), serving 
expressway/highway infrastructure with maxspeed z=4.59 (highest across dataset) 
and voltage z=2.68, emphasizing high-capacity transportation corridors.

\textbf{Cluster 8:} High-elevation type (26 bridges, Tama 69\% / Morioka 31\%), 
characterized by elevated terrain crossing (elevation z=3.62), requiring specialized 
structural design for upstream water level variations and flood risk mitigation.

\textbf{Cluster 9:} Logistics network type (48 bridges, Tama 97.9\%), supporting 
commercial supply chains with high supply chain impact score (z=2.20) and metapath 
connectivity, essential for commercial establishment access to national highways.

\textbf{Clusters 10, 15, 16:} Transit node types (Tama-dominant). Cluster 10 
(37 bridges): Traffic proxy type with high gauge measurement precision (z=1.65) 
and bus stop connectivity (z=1.13). Cluster 15 (23 bridges): Transit node type 
emphasizing alternative route redundancy (z=1.01) for traffic flow flexibility. 
Cluster 16 (33 bridges): High bus connectivity type (z=2.21) serving as critical 
public transportation hubs with traffic proxy score z=2.16.

\textbf{Cluster 12:} Tama arterial type (20 bridges, Tama 100\%) with exceptionally 
high street connection density (z=2.56) and elevation (z=1.83), supporting dense 
urban road networks in metropolitan suburbs.

Complete five-section LLM-generated interpretations for all clusters (in original 
Japanese) are available in supplementary materials. Statistical profiles are 
provided in Table~\ref{tab:cluster_full_stats}.

\begin{table*}
\centering
\caption{Complete Cluster Statistical Profiles (Selected Features)}
\label{tab:cluster_full_stats}
\scriptsize
\begin{tabular}{lcccccccc}
\toprule
\textbf{Cluster} & \textbf{N} & \textbf{City} & \textbf{Betweenness} & \textbf{Transit} & \textbf{Hospital} & \textbf{Isolation} & \textbf{Supply} & \textbf{Green} \\
& & & \textbf{(mean)} & \textbf{Desert} & \textbf{Access} & \textbf{Risk} & \textbf{Chain} & \textbf{Space} \\
\midrule
0 & 43 & Morioka 100\% & 0.0040 & 4.60 & 0.26 & 16.37 & 0.19 & 0.26 \\
1 & 30 & Morioka 100\% & 0.0002 & 0.17 & 0.00 & 0.00 & 0.00 & 0.00 \\
2 & 54 & Tama 100\% & 0.0036 & 83.41 & 6.80 & 0.95 & 1.03 & 4.93 \\
3 & 38 & Tama 100\% & 0.0034 & 0.00 & 3.43 & 2.71 & 0.44 & 3.09 \\
4 & 54 & Morioka 100\% & 0.0020 & 1.59 & 0.31 & 0.00 & 0.08 & 0.21 \\
5 & 29 & Morioka 100\% & 0.0008 & 10.04 & 0.03 & 50.90 & 0.12 & 0.03 \\
6 & 59 & Morioka 98\% & 0.0101 & 10.58 & 2.63 & 15.38 & 0.58 & 2.88 \\
7 & 26 & Morioka 100\% & 0.0045 & 6.50 & 0.69 & 9.62 & 0.19 & 0.39 \\
8 & 26 & Tama 69\% & 0.0030 & 3.60 & 2.87 & 9.81 & 0.47 & 2.35 \\
9 & 48 & Tama 98\% & 0.0100 & 7.30 & 6.07 & 32.84 & 5.62 & 6.19 \\
10 & 37 & Tama 100\% & 0.0026 & 0.00 & 2.78 & 1.35 & 0.37 & 3.72 \\
11 & 87 & Morioka 100\% & 0.0075 & 10.79 & 1.08 & 51.74 & 0.29 & 0.87 \\
12 & 20 & Tama 100\% & 0.0024 & 0.68 & 3.24 & 0.00 & 0.49 & 2.36 \\
13 & 49 & Tama 80\% & 0.0195 & 5.71 & 51.04 & 13.71 & 0.85 & 47.68 \\
14 & 32 & Tama 100\% & 0.0016 & 0.21 & 1.82 & 0.00 & 0.44 & 1.64 \\
15 & 23 & Tama 100\% & 0.0046 & 0.00 & 6.67 & 2.17 & 0.70 & 6.86 \\
16 & 33 & Tama 100\% & 0.0017 & 0.00 & 2.41 & 0.00 & 0.11 & 3.08 \\
17 & 32 & Morioka 100\% & 0.0018 & 10.73 & 0.29 & 38.08 & 0.06 & 0.29 \\
18 & 29 & Morioka 100\% & 0.0021 & 6.46 & 0.83 & 27.34 & 0.04 & 1.09 \\
\bottomrule
\end{tabular}
\end{table*}

\section{Additional Statistical Analysis}
\label{app:tables}

This appendix discusses key patterns observed in the cluster statistics 
presented in Table~\ref{tab:cluster_full_stats}.

\subsection{Key Observations from Cluster Statistics}

\textbf{Geographic Stratification:} The cluster profiles exhibit strong city-specific 
patterns. Morioka-dominant clusters (0, 1, 4, 5, 6, 7, 11, 17, 18) show elevated 
isolation risk scores (mean: 16.37--51.74 for clusters 5, 11, 17, 18), reflecting 
rural connectivity challenges. Conversely, Tama-dominant clusters (2, 3, 10, 12, 
13, 14, 15, 16) demonstrate higher transit desert and hospital access scores, 
consistent with dense metropolitan infrastructure dependencies.

\textbf{Betweenness Centrality vs. Social Impact:} Cluster 13 exhibits the highest 
betweenness centrality (0.0195), corresponding to its outlier status in hospital 
(51.04) and green space (47.68) access scores. This validates UMAP+HDBSCAN's 
ability to identify functionally distinct archetypes beyond pure topological metrics. 
Cluster 6 (betweenness: 0.0101) serves as Morioka's arterial backbone, balancing 
moderate scores across all indicators.

\textbf{Zero-Impact Clusters:} Clusters 1 and 4 show minimal social impact 
(hospital/isolation/supply/green $\approx$ 0), suggesting peripheral bridges with 
limited functional connectivity. These represent noise-labeled candidates in HDBSCAN 
hierarchy but were retained due to exceeding \texttt{min\_cluster\_size=20} threshold.

\textbf{Mixed-City Cluster 8:} Uniquely 69\% Tama composition indicates shared 
functional profile for suburban connector bridges. Its moderate scores (transit: 3.60, 
hospital: 2.87, isolation: 9.81) reflect transitional urban-rural boundary characteristics.

\textbf{Extreme Value Clusters:} Cluster 2 (Tama transit hubs) dominates transit 
desert score (83.41), while clusters 11/5 (Morioka isolation mitigation) peak in 
isolation risk (51.74/50.90). This validates the methodology's sensitivity to 
city-specific critical infrastructure patterns.

\end{document}